\documentclass[10pt,twocolumn,letterpaper]{article}

\usepackage{iccv}
\usepackage{times}
\usepackage{epsfig}
\usepackage{graphicx}
\usepackage{amsmath}
\usepackage{amssymb}
\usepackage[normalem]{ulem}
\useunder{\uline}{\ul}{}
\usepackage{caption}
\usepackage{multirow}
\usepackage{subfigure}
\usepackage{makecell}
\usepackage{authblk}
\usepackage[table,xcdraw]{xcolor}
\usepackage[pagebackref=true,breaklinks=true,letterpaper=true,colorlinks,bookmarks=false]{hyperref}

\iccvfinalcopy 

\def\iccvPaperID{8839} 

\ificcvfinal\fi

\begin{document}
\definecolor{mygray}{gray}{.9}

\title{ADNet: Lane Shape Prediction via Anchor Decomposition}

\author{Lingyu Xiao$^{~\dagger}$, Xiang Li$^{~\ddagger}$, Sen Yang$^{~\dagger}$, Wankou Yang$^{~\dagger}$\thanks{Corresponding author}\\
    $^{\dagger}$~School of Automation, Southeast University, China\\
    $^{\ddagger}$~IMPlus@PCALab, VCIP, CS, Nankai University, China\\
    {\tt\small lyhsiao@seu.edu.cn, xiang.li.implus@njust.edu.cn, \{yangsenius, wkyang\}@seu.edu.cn}\\
}

\maketitle
\ificcvfinal\thispagestyle{empty}\fi

\begin{abstract}
In this paper, we revisit the limitations of anchor-based lane detection methods, which have predominantly focused on fixed anchors that stem from the edges of the image, disregarding their versatility and quality. To overcome the inflexibility of anchors, we decompose them into learning the heat map of starting points and their associated directions. This decomposition removes the limitations on the starting point of anchors, making our algorithm adaptable to different lane types in various datasets. To enhance the quality of anchors, we introduce the Large Kernel Attention (LKA) for Feature Pyramid Network (FPN). This significantly increases the receptive field, which is crucial in capturing the sufficient context as lane lines typically run throughout the entire image. We have named our proposed system the Anchor Decomposition Network (ADNet). Additionally, we propose the General Lane IoU (GLIoU) loss, which significantly improves the performance of ADNet in complex scenarios. Experimental results on three widely used lane detection benchmarks, VIL-100, CULane, and TuSimple, demonstrate that our approach outperforms the state-of-the-art methods on VIL-100 and exhibits competitive accuracy on CULane and TuSimple. Code and models will be released on \url{https://github.com/Sephirex-X/ADNet}.
\end{abstract}

\section{Introduction}\label{chp.introduction}

Recently, the utilisation of artificial intelligence technology for the field of autonomous driving has drawn large attention from academia and industry. As a crucial part of autonomous driving system, Advance Driver Assistance System (ADAS) requires vehicles to respond timely and accurately to changes in the environment. Lane line is a vital part of the vehicle sensing the environment, as the ADAS needs the shape of lane lines to keep the vehicle on track.
\begin{figure}[t]

\subfigure[Annotations]{
\begin{minipage}[t]{0.32\linewidth}
\centering
\includegraphics[width=1.05in]{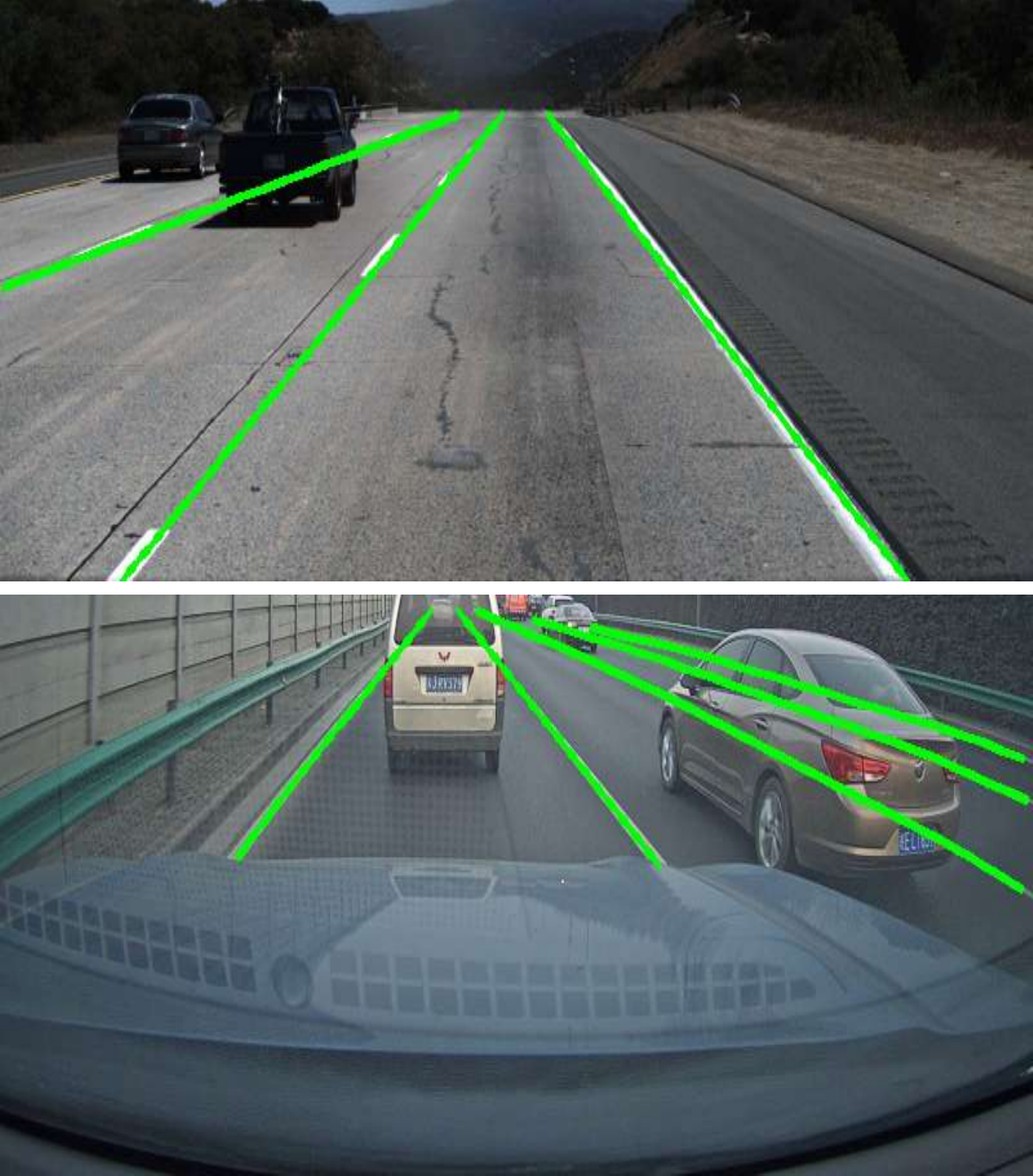}
\label{introduction_a}
\vspace{-20mm}
\end{minipage}%

}%
\subfigure[Traditional]{
\begin{minipage}[t]{0.32\linewidth}
\centering
\includegraphics[width=1.05in]{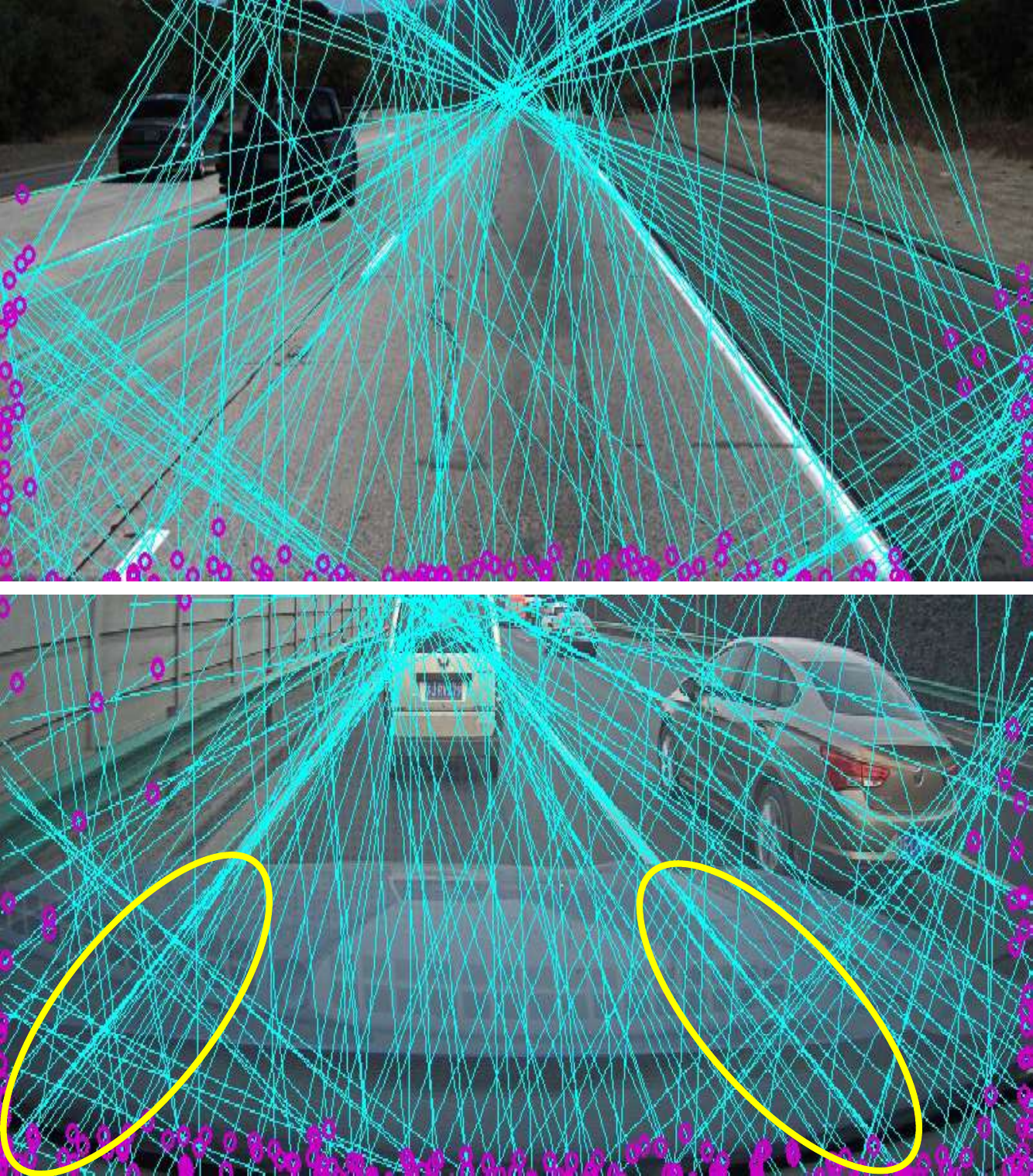}
\label{introduction_b}
\vspace{-20mm}
\end{minipage}%
}%
\subfigure[Ours]{
\begin{minipage}[t]{0.32\linewidth}
\centering
\includegraphics[width=1.05in]{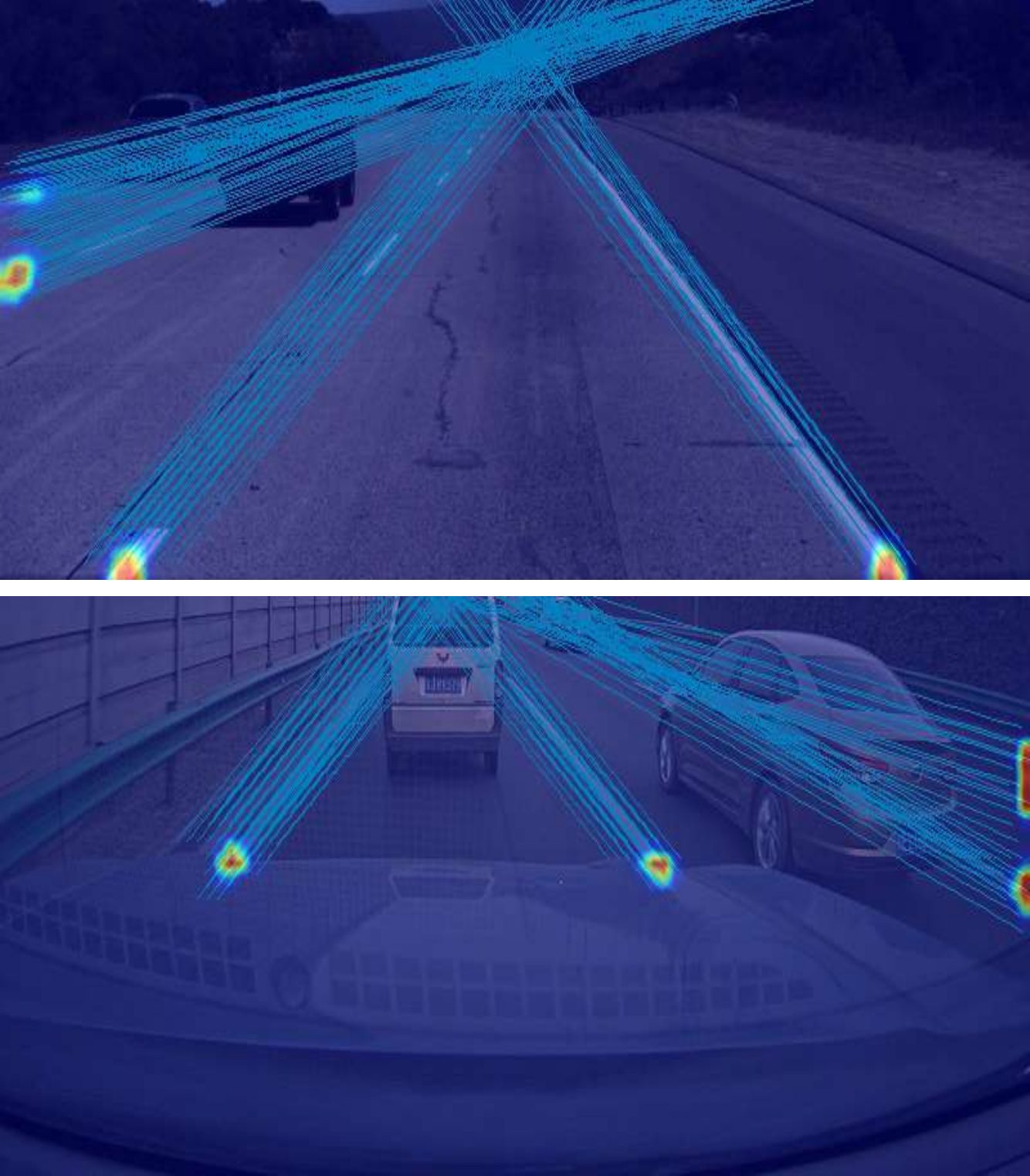}
\label{introduction_c}
\vspace{-20mm}
\end{minipage}%
}%
\vspace{-3mm}
\caption{Illustration of different dynamic anchor proposal methods. (a) illustrates two common lane prediction scenarios. In the first row, the lane lines originate from the edges of the image, while in the second row, the lane lines can emanate from any location within the image. (b) proposes anchor \textbf{dispersedly}~\cite{Clrnet}, resulting in low anchor quality. This anchor proposal method is adequate on first row scenarios but oversimplistic on the second (emphasised by \textcolor{yellow}{yellow} oval). The points and lines represent start points and anchors respectively. 
(c) we propose anchor \textbf{concentratively}, possible start points are shown on activation map, anchors can merge from the whole image, which ensures anchor quality and flexibility.
}
\label{Dynamic ank}
\vspace{-5mm}
\end{figure}

With the advancement of CNN, recent studies on lane shape detection have made great progress on either accuracy~\cite{Laneatt,GANnet,Clrnet} or real-time performance~\cite{UFLD,UFLDv2,ConditionLane}. Anchor-based methods have shown superior accuracy and efficiency compared to other methods on popular benchmarks such as~\cite{SCNN,LLamas,Tusimple}. However there are still challenges to the wide application of anchor-based methods. 

The first issue is the flexibility of anchors. Previous anchor-based methods~\cite{Laneatt,LineCNN,Clrnet,UFLD} have posited that lane lines can only originate from the three edges of an image (left, bottom, right). While this assumption leverages prior information of lane lines to achieve favourable accuracy and speed, it is oversimplistic as lane lines do not always start from the three edges due to the obstructions such as vehicles in adjacent lanes or a vehicle's front hood (shown on Figure \ref{introduction_a}). 

Another problem is the low quality of anchors. Anchor-based methods~\cite{LineCNN,Laneatt} usually employ an approach of fixed anchors, while recent method~\cite{Clrnet} adopt a dynamic approach with dispersed anchor prediction. This dispersed prediction (shown on Figure \ref{introduction_b}) is possibly unreliable when the camera resolution is varying, particularly in cases where lane lines do not start from the edges.
Additionally, the inherent physical characteristics of lane lines, such as their slenderness and continuity, presenting significant challenges in capturing their geometric features. {However, most existing approaches are limited in small kernel sizes which present an obstacle to accurately capture the whole feature descriptors of lane lines.}

{In this paper, to address these problems, we introduce Anchor Decomposition Network (ADNet). Specifically, to make anchors flexible, we propose Start Point Generate Unit (SPGU) {which decomposes them into predicting the position of the start points and its associated direction on a global scale by the probability map (heat map). }
To enhance anchors' quality, we realise the crucial role of large receptive in capturing slender and continue lane lines. Therefore we 
{introduce a Large Kernel Attention (LKA) module and integrate it with the Feature Pyramid Network (FPN).} Since we predict anchors in a {concentrative} way (shown in Figure \ref{introduction_c}), the location is invariant to the density of validated pixels and thus the anchors' quality and flexibility can be mutually guaranteed. } 

We conduct extensive experiments on three lane detection benchmarks: VIL-100~\cite{VIL-100}, CULane~\cite{SCNN} and TuSimple~\cite{Tusimple}. Comparing along with state-of-art methods, our approach shows excellent performance on all datasets. In particular, on VIL-100 dataset where segmentation-based methods always perform superior to the anchor/keypoint-based counterparts, our framework outperforms all existing state-of-art methods, making the anchor-based method a more generalised pipeline. The main contributions of this paper are summarised as :
\begin{itemize}
    \item {We emphasise the importance of anchor flexibility for the anchor-based approaches by explicitly decomposed learning of the heat map of starting points and their associated directions. The decomposition makes our algorithm adaptable to different lane types in more scenarios.}
    \item {To our best knowledge}, we are the first to {investigate the effectiveness of  Large Kernel mechanism}
    on lane detection task to guarantee the anchor quality, {as the lane lines usually cover the entire image which often require significantly large context to be accurately captured.} 
    \item We rethink the limitations of LIoU loss and propose our own General Line IoU (GLIoU) loss tailor for anchor-based lane detection method on complex scenarios. {Furthermore, we utilise the explicit physical modelling by anchor decomposition to guide the learning of kernel offsets in the proposed Adaptive Lane Aware Unit (ALAU).}
    \item Experiments on main benchmarks show excellent trade-offs on performance and speed compared with SOTA methods, outperforming all recent methods on VIL-100 dataset.

\end{itemize}

\begin{figure*}[t]
\begin{center}
 \includegraphics[width=1\textwidth]{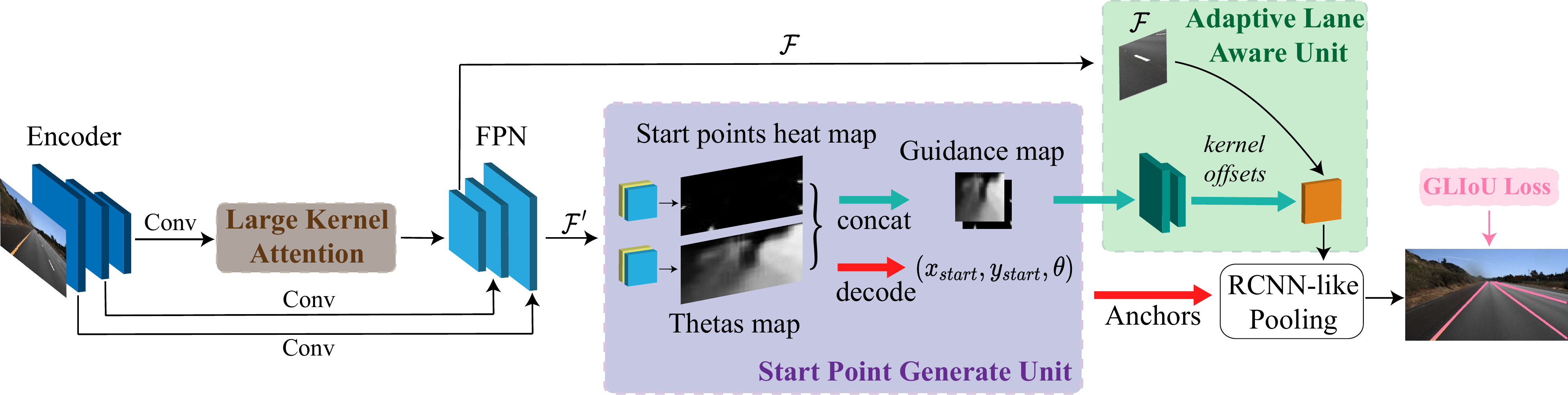}  
\end{center}
\vspace{-5mm}
    \caption{Overview of our ADNet. Lane context first extracted by the encoder and enhanced by FPN embedded with Large Kernel Attention (LKA), which plants after FPN’s lateral layer to reduce computation cost. Then, low-level context $\mathcal{F}'$ is delivered into Start Point Generate Unit (SPGU) to generate start point guided anchors and guidance map, while high-level context $\mathcal{F}$ is further aggregated through Adaptive Lane Aware Unit (ALAU) with the help of the auxiliary guidance map. After pooling, we optimise lane lines via General Lane IoU loss. }
\label{overall_structure}
\vspace{-5mm}
\end{figure*}
\section{Related Work}\label{chp.related_work}

\subsection{Segmentation-based methods}
In segmentation-based methods, the task of identifying lane lines has been converted to a per-pixel prediction task.~\cite{SCNN} first introduces a spatial mechanism passing messages between pixels row-wise and column-wise that fails to perform in real time.~\cite{RESA} further proposes a recurrent aggregator fully utilised lane shape priors to obtain better performance. On~\cite{LaneAF}, additional affinity fields are predicted simultaneously with the binary segmentation map, which is used in the decoder to cluster lane pixels. Segmentation-based method can achieve high accuracy when lane lines are visible, but it's unstable in complex traffic scenarios and inefficient.

\subsection{ Anchor-based methods}

Anchor-based \& detection-based methods define lane lines in a similar way. They divide an image into slices or cells, and then convert the lane detection task into either offsets' regression on each slice or a row-wise classification task.~\cite{UFLD} first predicts lane lines via a simple linear layer using row-wise classification.~\cite{LineCNN} improves the representation of lane lines by converting cell representation into anchor representation, and identifying lane shape through regression of the offsets on every slice between anchors and ground truth.~\cite{Laneatt} further enhances this formulation by adding anchor-based pooling and a lane attention mechanism to it.~\cite{UFLDv2} proposes a hybrid anchor system to improve the performance of UFLD.~\cite{ConditionLane} proposes a conditional convolution and RIM migration to solve the instance-level discrimination problem on lane detection.~\cite{Clrnet} develops ROIGather to fuse lane context from different layers and, for the first time, changes the anchor-based formulation into an anchor-free manner, achieving state-of-the-art performance on multiple benchmarks.

Anchor-based and detection-based methods heavily rely on the position of anchors. On one hand, this can bring higher accuracy since anchors contain prior information on lane lines. On the other hand, these inherent properties lead to some shortcomings, such as the starting point of the anchor may not always be located on the three edges of the image, limiting its application.

\subsection{Keypoint-based methods}
Keypoint-based methods treat lane lines’ prediction as a key point estimation task. Usually, the algorithm will first predict all the possible key points that most likely belong to lane lines, and follow up with a post-process of assigning different points to different lanes.~\cite{PINet} predicts key points on lane lines and distinguishes each instance by embedding features of predicted points.~\cite{FOLOLane} predicts local key points in a bottom-up manner and refiners key points’ location via its offsets between adjacent points.~\cite{GANnet} clusters points via offsets between key points and start points, and a modified deformable convolution network~\cite{DCN} to extract holistic lane features. Lane instances are predicted by keypoint-based methods via low-efficient post-processing of key points from the heat map, moreover, the accuracy of the algorithm highly relies on the resolution of the input image, together with time-consuming post-processing, making keypoint-based methods hard to strike a balance between latency and accuracy.


\section{Approach}

The overall structure is illustrated in Figure \ref{overall_structure}. 
{Our algorithm contains} four parts: Start Point Generate Unit (SPGU), Adaptive Lane Aware Unit (ALAU), General Lane IoU loss and Large Kernel Attention (LKA).

\subsection{Start point generate unit}

\textbf{Motivation.} On lane shape detection tasks, predefined anchors have direct affection toward the anchor-based \& detection-based method~\cite{Laneatt,LineCNN}. Literature like~\cite{Clrnet,SPNet} perform in an anchor-free manner but they work under the assumption that lane line rays from three edges of the image, therefore, limiting its application.

\textbf{Structure.} The ultimate goal is to form an anchor by predicting the start point location and theta given downsampled feature map $\mathcal{F'} \in \mathbb{R} ^{H_f'\times W_f'}$, which can be formulated as $p(x_{start},y_{start},\theta | \mathcal{F'})$ .
Like most of the Keypoint-based detection framework~\cite{Cornernet,Extremenet,Centernet,reppoints}, we aim to predict the start point for each lane on the image by estimating the possibility of a start point on a certain region of the downsampled heat map. {Additionally, we observe that the theta of the anchor is closely associated with the start point in terms of spatial relation~\cite{Guidedanchoring}}, {according to the Bayes' theorem,} its location and shape can be {decomposed} as: 
\begin{equation}
\resizebox{\linewidth}{!}{$p(x_{start},y_{start},\theta | \mathcal{F'}) = p(x_{start},y_{start}|\mathcal{F'})p(\theta |x_{start},y_{start},\mathcal{F'}).$}
\end{equation}

During the training phase, we generate a supervision heat map by adding a non-normalised Gaussian kernel to each ground truth start points:
\begin{equation}
\resizebox{\linewidth}{!}{$H^{pts}_{gt}(x,y) = exp(-\frac{(x-x^{start}_{gt})^2+(y-y^{start}_{gt})^2}{2\sigma^2}) \ , \ H_{gt} \in \mathbb{R} ^{H_f'\times W_f'},$}
\end{equation}
     $x,y$ is the coordinate of pixels on $H^{pts}_{gt}$; $x^{start}_{gt},y^{start}_{gt}$  is ground truth start point’s coordinate; $\sigma$ is a hyperparameter. Then supervision for theta map $H^{\theta}_{gt}(x,y)$ can be formulated as:
\begin{equation}
\resizebox{\linewidth}{!}{$H^{\theta}_{gt}(x,y) = index(H^{pts}_{gt}(x,y) > t_\theta) \cdot \theta(x^{start}_{gt},y^{start}_{gt}).$}
\end{equation}

We can interpret this as follows: if the probability of start points in a particular region is greater than $t_\theta$, we consider that region to share the same $\theta$ as the ground truth start point. Unlike a strict assumption of one-point-one-theta, our approach expands the potential occurrence area of proposal anchors, providing a wealth of high-quality anchors for regression. This allows neural networks to determine the best fit under different conditions, without needing to add an additional loss to compensate for the offset between points on the heat map and the original image due to downsampling~\cite{GANnet}.

We modified~\cite{Focalloss} to meet the imbalance between start point regions and the {non-start point regions}:
 \begin{equation}
\resizebox{\linewidth}{!}{$     \mathcal{L}_{hm} = \frac{-1}{H'_f \times W'_f}\sum_{xy}
\begin{cases}
(1-H^{pts}_{pred})^\alpha log(H^{pts}_{gt}) & H^{pts}_{gt}=1 \\
(1-H^{pts}_{gt})^\beta (H^{pts}_{pred})^\alpha log(1-H^{pts}_{pred}) & otherwise
\end{cases},$}
\label{eq.hm_loss}
\end{equation}
 $\alpha$ and $\beta$ are hyperparameters of focal loss. Similarly, we modified L1 loss for theta map over the whole feature map:
 \begin{equation}
     \mathcal{L}_\theta=\frac{1}{H'_f\times W'_f}\sum_{xy}|H^{\theta}_{pred}-H^{\theta}_{gt}|.
\label{eq.theta_loss}
 \end{equation}

It is noteworthy that the calculation of theta map loss in the non-start point region is unnecessary due to the uncertainty of their theta values.
\subsection{Large kernel attention}

\begin{figure}[t]
\subfigure[]{
\begin{minipage}[t]{0.5\linewidth}
\centering
\includegraphics[width=1in]{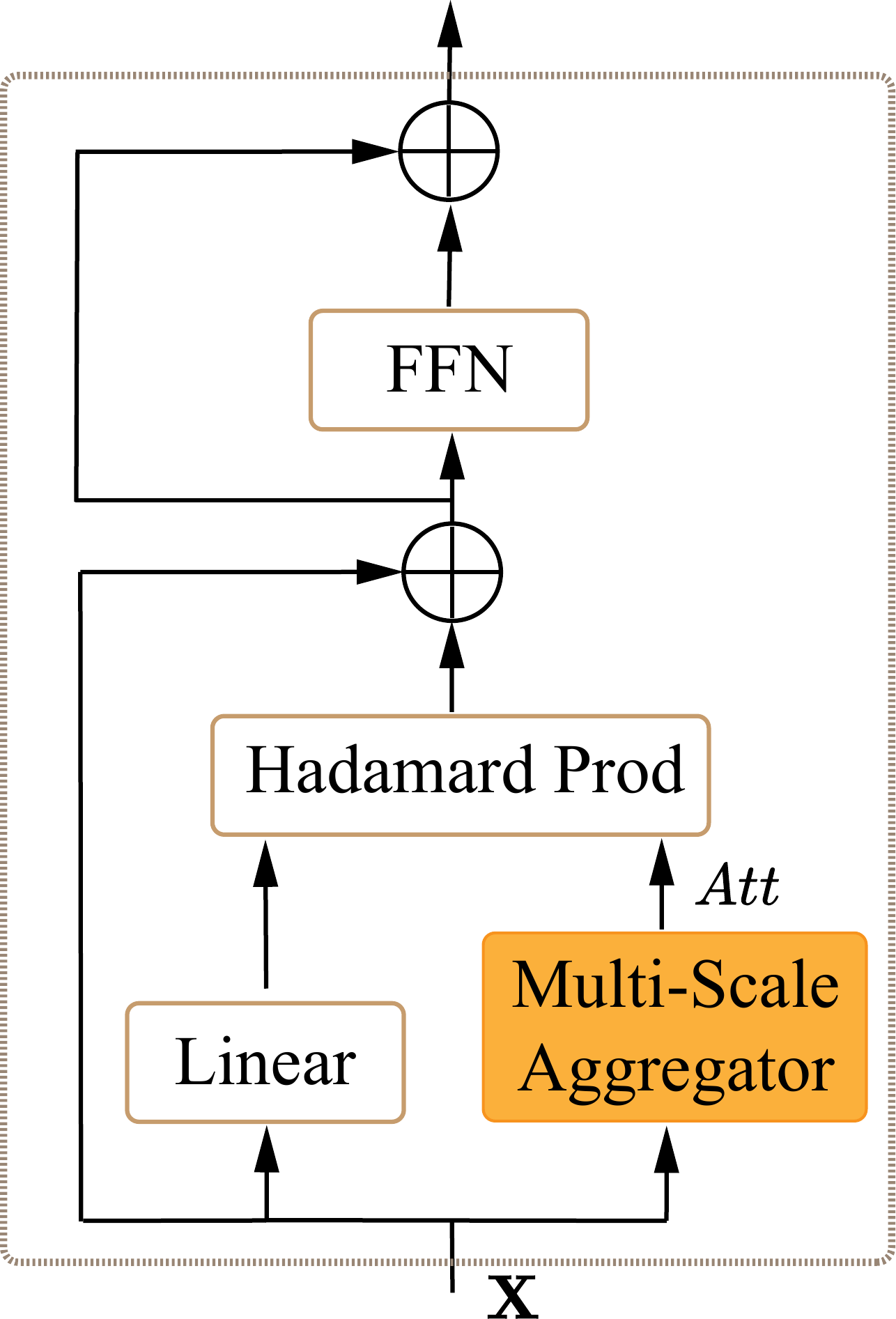}
\label{LKA-B}
\end{minipage}%
}%
\subfigure[]{
\begin{minipage}[t]{0.5\linewidth}
\centering
\includegraphics[width=1.5in]{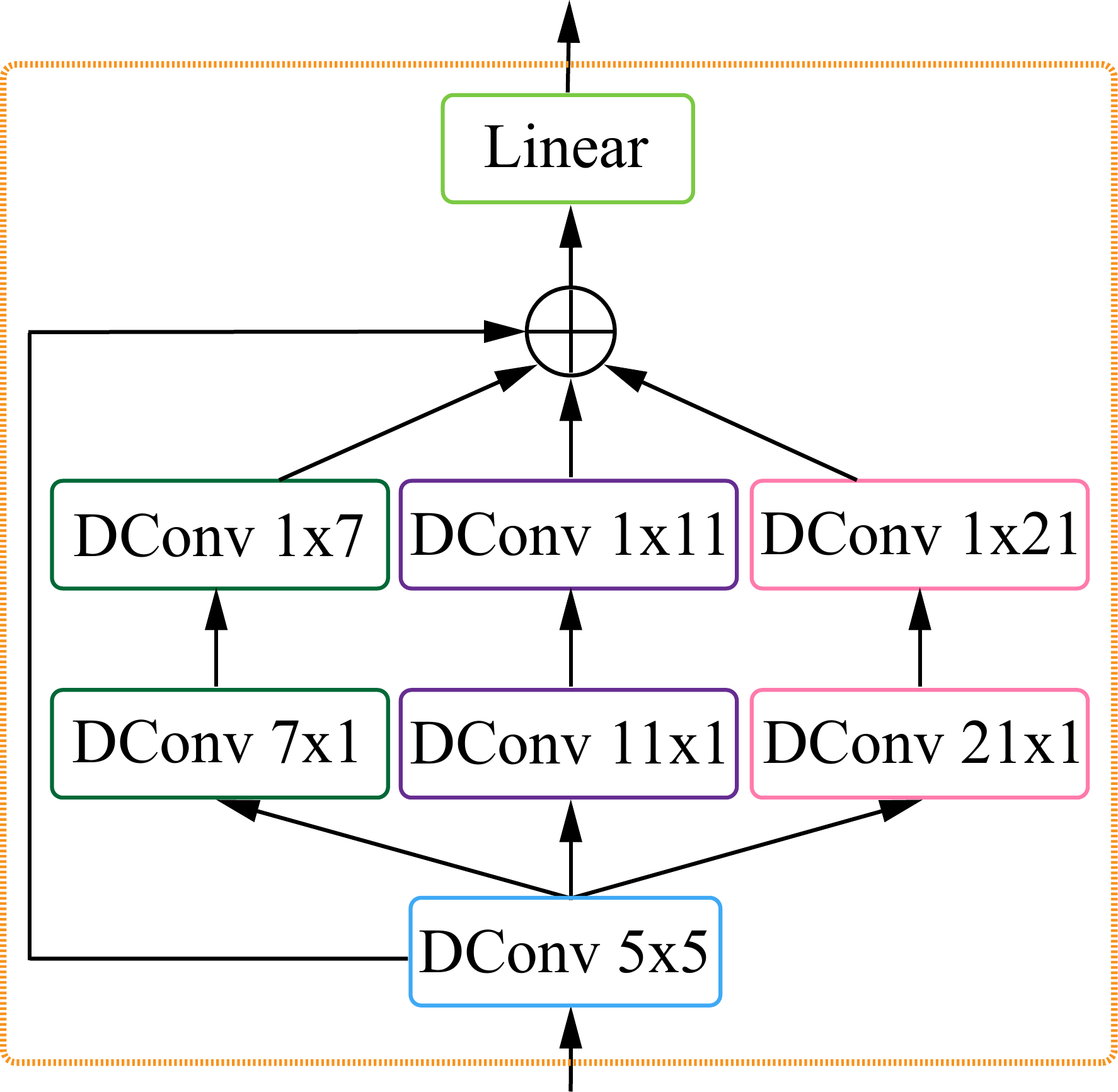}
\label{LKA-C}
\end{minipage}%
}%
\vspace{-3mm}
\caption{Illustration of LKA. LKA can be seen as the combination of (a) attention mechanism and (b) Multi-Scale Aggregator (MSA) .}
\label{LKA}
\vspace{-4mm}
\end{figure}

\textbf{Motivation.} 
In recent literature~\cite{Revisitinglargekernel,Repvgg}, it has been observed that the performance of ConvNet is restricted when the kernel size exceeds $7\times7$, thereby limiting the potential benefits of mixed Transformer architecture for downstream tasks that require a large receptive field. Building on the work of~\cite{conv2former,segnetxt}, we devise a Large Kernel Attention (LKA) module integrated with Feature Pyramid Network (FPN) specifically for lane detection.

\textbf{Design.} In Figure \ref{overall_structure}, it can be seen that our LKA module is placed after the lateral layer of FPN to minimise computation cost. Unlike generating a similarity score $Att$ between the query and value outputs, we employ Multi-Scale Aggregator (MSA) to quantify the correlation among input tokens. The mathematical expression of our approach depicted in Figure \ref{LKA} can be represented as follows:
\begin{align}
 & Att = \mathbf{W}_1(\sum_{i=0}^3 MultiCh_i(DConv_{5\times 5}(\mathbf{X}))), \\
 & \mathbf{Z}_1=Att \odot (\mathbf{W}_2\mathbf{X}) + \mathbf{X}, \label{eq.hadamard}  \\ 
 & \mathbf{Z} = FFN(\mathbf{Z}_1) + \mathbf{Z}_1.
\end{align}

In Figure \ref{LKA-C}, the four feed-forward paths are denoted as ${MultiCh_i}$ and are distinguished by different colours, where $MultiCh_0$ corresponds to an identical forward path. Instead of using a $7\times7$ depth-wise convolution, strip-like convolutions are more effective in identifying lane lines while reducing computation cost. The linear layer is represented by $\mathbf{W}_i$. As suggested in~\cite{conv2former}, we use Hadamard product (denoted as $\odot$ in Eq.~\eqref{eq.hadamard}) instead of matrix product to leverage the advantages of large kernels in MSA.

\subsection{General Lane IoU loss}\label{chp.gliou}

\begin{figure}[t]
\centering
\includegraphics[width=0.8\linewidth]{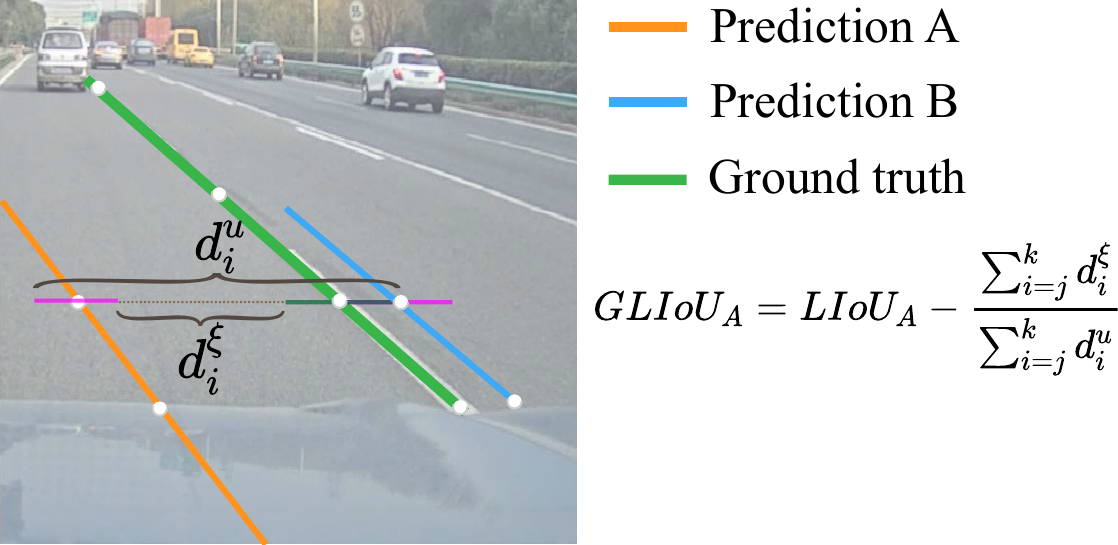}
\caption{Illustration of GLIoU. A special scenario that LIoU fails to address properly. We have extended each point on the slice to form lane segments with a certain width $e$, and it is evident that the L1 distance between the two segments is significantly larger than what can be captured by LIoU ($LIoU_A=-0.5, LIoU_B=0.5$, $distanceL_1^A=6e$, $distanceL_1^B=e$).
}
\vspace{-3mm}
\label{GLIoU}
\end{figure}

\textbf{Motivation.} 
{Recently, LIoU~\cite{Clrnet} loss has been proposed to address the problem that the lane shape information on the anchor-based method is considered to be independent for each point when applying L1 loss. Although LIoU loss incorporates information on lane shape into a normalized metric that is invariant to scale, it may not be suitable for infrequent scenarios.} Figure \ref{GLIoU} depicts a typical scenario that exposes the limitations of LIoU. As shown in Figure \ref{GLIoU}, the LIoU values for \textit{Prediction A-Gt} and \textit{Prediction B-Gt} are -0.5 and 0.5, respectively, whereas the L1 distance gap between the two is significantly larger than what LIoU can capture ($distanceL_1^A=6e$, $distanceL_1^B=e$). In other words, \textit{Prediction A} is substantially worse than \textit{Prediction B} according to the L1 distance metric, yet LIoU fails to account for this relationship.

\textbf{Design.} {To overcome this limitation, we propose General Lane IoU (GLIoU), which can be considered as a generalisation of LIoU, where an additional penalty term is incorporated to highlight the spatial relationship between two lanes that do not overlap.} Similar to LIoU, we begin by extending each point to form lane segments with a certain width $e$ and computing the intersection over union ratio as usual. Then, we calculate the L1 distance between each pair of extended segments to obtain a gap distance $d^\xi_i$:
 \begin{equation}
     d^\xi_i = ReLU(d^u_i-4e).
     \label{eq.di}
 \end{equation}

Using LIoU subtract the ratio of gap distance to the union, which is illustrated as follows:
\begin{equation}
    GLIoU = \frac{\sum_{i=j}^kd^o_i-ReLU(d^u_i-4e)}{\sum_{i=j}^kd^u_i},
    \label{eq.GLIoU}
\end{equation}
where $j$ represents the index of the first validated point. On Eq.~\eqref{eq.di} and Eq.~\eqref{eq.GLIoU}, $d^u_i$ and $d^o_i$ is defined as introduced in~\cite{Clrnet}. If the prediction overlaps with the ground truth, GLIoU degenerates into LIoU. However, if the prediction does not overlap with the ground truth, we introduce an extra penalty term $\frac{\sum_{i=j}^kd^\xi_i}{\sum_{i=j}^kd^u_i}$ to more accurately capture the L1 distance while still considering the lane as a unified entity.

The GLIoU loss can be defined as:
\begin{equation}
    \mathcal{L}_{GLIoU} = 1-GLIoU.
\label{eq.gliou}
\end{equation}

The domain of GLIoU is $(-2,1]$, when the predicted lane perfectly matches the ground truth, GLIoU equals 1. Although the lower bound of GLIoU is -2, which may seem asymmetric, it is a more appropriate choice for predicting lane shapes since, in most cases, the prediction and ground truth do not perfectly align. Rather than emphasising the overlapped section, the GLIoU loss focuses on improving the poorly overlapped section.

\subsection{Adaptive lane aware unit} \label{chp.alau}

\textbf{Motivation.} 
The use of traditional convolution networks for lane detection may not be optimal, as they operate on a fixed grid that does not align well with the irregular shape of lane lines. Although the Deformable Convolution Network (DCN)~\cite{DCN} has found extensive application in object detection, its potential for lane detection has not been fully explored~\cite{GANnet}. Directly applying DCN to lane detection is challenging, as it is not feasible to learn kernel offsets from high-context lane features. {Instead, we observe that start points and their associated thetas can be regarded as an effective guidance to predict kernel offsets due to their explicit physical modelling.}

\begin{figure}[t]
\begin{center}
\includegraphics[width=0.7\linewidth]{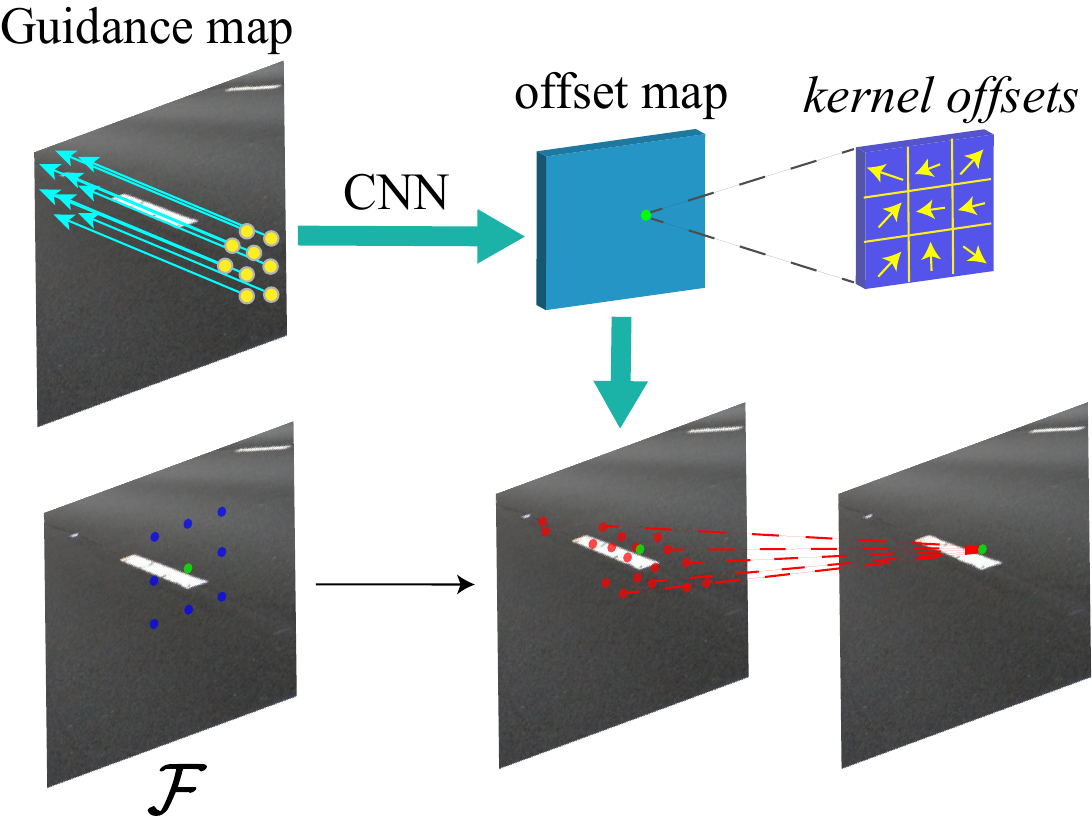}
\end{center}
\vspace{-3mm}
    \caption{Illustration of ALAU. We utilise the SPGU-generated guidance map to predict kernel offsets and embed them with deformable convolution to gather lane context. In the following image, the green dot represents the activation unit, the yellow arrow indicates the offset learned from the guidance map, and the red dot denotes the sampling location  ($9\times2\times1=18$ on each image) in a single $3\times3$ kernel level. The deformable group is set to 2.}
\label{alau}
\vspace{-3mm}
\end{figure}

\textbf{Structure.} 
Once we have obtained the start points coordinates and theta values that are spatially related using SPGU, we encode the thetas heat map and start points heat map with dense lane information, which can be represented as $\Theta_{xy}=\{\theta_1,\theta_2,...,\theta_N \}$ and $P_{xy}=\{p_1,p_2,...,p_N\}$ respectively, as shown in Figure \ref{alau}. For instance, we can abstract the task of predicting a set of kernel offsets on one activation unit (green dot) as follows:
\begin{align}
& \Delta K_{xy} = \phi(\vec{v} \cdot pts)=\phi(x,y,\theta), \label{eq.delta}  \\
& \mathcal{S} = \{ (-1,-1),(0,-1),...,(0,1),(1,1) \}, \\ 
& \Delta K_{xy} = \{ \Delta k_{xy}^i|i =1,...,|\mathcal{S}| \}. 
\end{align}

Let $\mathcal{S}$ denote the grid defined by the receptive field size and dilation, and let $pts$ denote the position of the activation unit, with coordinates $x$ and $y$. Let $\vec{v}$ be a unit vector parallel to the spatially nearest anchor, with $\theta$ denoting the anchor's theta value. Let $\Delta K_{xy}$ be the kernel offsets on $pts$, where $ks$ is the kernel size and $\phi$ is a non-linear function. Since the theta value in non-start point regions is uncertain and the anchors' direction is mutually related, SPGU will automatically learn the theta value that has the highest probability for the region, ensuring the existence of the function $f_{v}: \Theta_{xy} \to \vec{v}$. Since the start point and its theta value are spatially related, the function can be further expressed as $f_{v}: (\Theta_{xy},P_{xy}) \to \vec{v}$. Therefore, Eq.~\eqref{eq.delta} can be rephrased as:
\begin{equation}
    \Delta K_{xy}= \phi(f_v( \Theta_{xy},P_{xy} )).
\end{equation}

We can simply use convolutional neural network to fit these functions. Following~\cite{DCN} we integrated kernel offsets with deformable convolution to adaptively extract context of activation unit, which can be expressed as:
\begin{equation}
    \hat{\mathcal{F}}(pts) = \sum_{pts^i_{xy} \in \mathcal{S}} w(pts_{xy}^i) \cdot \mathcal{F}(pts+pts_{xy}^i+\Delta k_{xy}^i),
\end{equation}
where $w(pts_{xy}^i)$ is the weights of convolution. In Figure \ref{alau}, the green dot, blue dots, yellow arrows, and red dots correspond to $pts$, $pts+pts^i_{xy}$, $\Delta k^i_{xy}$, and $pts+pts^i_{xy}+\Delta k^i_{xy}$, respectively.

\subsection{Model training detail}
\textbf{Label assignment.} Since the algorithm works in an anchor-free style, assigning a positive label in a predefined anchor manner is not feasible. We follow ~\cite{Clrnet} to assign labels dynamically~\cite{yolox}. 

\textbf{Loss function.} The overall loss function can be written as:
 \begin{equation}
     \mathcal{L} = w_{reg}\mathcal{L}_{GLIoU} + w_{cls}\mathcal{L}_{cls} + w_{hm}\mathcal{L}_{hm} + w_{\theta}\mathcal{L}_{\theta }. 
\label{eq.loss}
 \end{equation}

The loss function comprises four components: $\mathcal{L}_{GLIoU}$ is the General Lane IoU loss (Eq.~\eqref{eq.gliou}) between proposals and ground truths; $\mathcal{L}_{cls}$ is focal loss~\cite{Focalloss} between proposals and ground truths; $\mathcal{L}_{hm}$ is modified focal loss (Eq.~\eqref{eq.hm_loss}) between start points heat map and ground truth; $\mathcal{L}_{\theta}$ is modified L1 loss (Eq.~\eqref{eq.theta_loss}) between thetas heat map and ground truth. The total loss is obtained by taking the weighted sum of each component.
\section{Experiments}
\subsection{Datasets and evaluation metric}
In this paper, we use three popular benchmarks: VIL-100, CULane and Tusimple.

\textbf{VIL-100~\cite{VIL-100}} is a recently released video instance lane detection dataset, that contains 10,000 frames. There are 10 scenarios in collection including multi-weather, multi-traffic scenes, day and night. The resolution of the image varies from $640 \times 368$ to $1920 \times 1080$ and lanes may locate in 8 different places, which challenges the algorithm.

\textbf{CULane~\cite{SCNN}} contains 88,880 images for training. The main scene is urban traffic, which also includes various scenery, such as daytime, night, crowded, fog, etc., making it a very challenging dataset. All annotated images are $1640 \times 590$ pixels in size.

\textbf{TuSimple~\cite{Tusimple}} consists of simple scenes where lane lines are easily identifiable. Each annotated image has a size of $1280 \times 720$ pixels, and contains a maximum of five lane lines.

\textbf{Evaluation metric.} There are two main evaluation metrics widely used in lane detection: F1 and Accuracy. F1 is defined as $F1 = \frac{2\times Precision \times Recall}{Precision + Recall}$. To evaluate IoU, lanes are extended with a width of 30 pixels~\cite{SCNN}, and predictions with IoU greater than a threshold are considered as true positive (TP). Accuracy (Acc) is defined as $Acc = \frac{\sum_{clip}C_{clip}}{\sum_{clip}S_{clip}}$, where $C_{clip}$ is the number of points within 20 pixels of ground truth per image, denoted as correct points; $S_{clip}$ is the total number of points within an image. A prediction is considered correct if it has more than 85\% of points noted as correct points. False Positive Rate (FPR) and False Negative Rate (FNR) are defined as $FPR = \frac{F_{pred}}{N_{pred}}$ and $FN=\frac{M_{pred}}{N_{gt}}$, respectively.
\subsection{Implementation details}
We employ Resnet~\cite{Resnet} pre-trained on ImageNet~\cite{ImageNet} as the backbone. 

For the TuSimple dataset, we utilise Adam~\cite{Adam} optimiser with an initial learning rate of 2e-5 per batch, and train for 150 epochs using the CosineLR~\cite{consineLR} learning rate decay strategy. The number of anchors is set to 100, and the hyperparameters for the supervision of the heat map are set to $\sigma = 2$ and $t_{\theta}=0.2$, respectively. The weights for the loss function in Eq.~\eqref{eq.loss} are set to $w_{reg} = w_{cls} = w_{hm} =10$, and $w_{\theta} = 1$. 

For the CULane dataset, we use AdamW~\cite{AdamW} optimiser and train for 15 epochs with 300 anchors. The loss function's weights are $w_{reg} = w_{cls} = 6$, $w_{hm} = 2$, and $w_{\theta} = 3$, while the hyperparameters for the heat map are $\sigma = 4$ and $t_{\theta}=0.5$. 

We use the same training settings for VIL-100 as TuSimple, except the training epoch is 80. 

During training and inference, we resize input images for all datasets to $800\times320$. The extend radius $e$ in GLIoU is set to 15. For FPS test on Table \ref{tb.culane} and Table \ref{tb.vil}, we set the batch size to 1 and forward model for 2000 times. All experiments are conducted on a single RTX3090.

\begin{table*}[]
\begin{center}
\caption{Comparison with state-of-art methods on CULane test set. `R18' stands for ResNet18, the rest can be analogised.}
\vspace{-2mm}
\label{tb.culane}
\resizebox{\linewidth}{!}
{
\begin{tabular}{cccccccccccc}
\hline
\textbf{Method}                          & \textbf{F1@50} $\uparrow$ & \textbf{FPS} $\uparrow$ & \textbf{Normal} $\uparrow$ & \textbf{Crowded } $\uparrow$ & \textbf{Dazzle} $\uparrow$ & \textbf{Shadow} $\uparrow$ & \textbf{No Line} $\uparrow$ & \textbf{Arrow} $\uparrow$ & \textbf{Curve} $\uparrow$ & \textbf{Cross} $\downarrow$ & \textbf{Night} $\uparrow$ \\ \hline
\textbf{Segmentation Based}              &                &              &                 &                  &                 &                 &                  &                &                &                &                \\ \cline{1-1}
SCNN-VGG16~\cite{SCNN}                               & 71.60          & 25             & 90.60           & 69.70            & 58.50           & 66.90           & 43.40            & 84.10          & 64.40          & 1990           & 66.10          \\
RESA-R50~\cite{RESA}                               & 75.30          & 65           & 92.10           & 73.10            & 69.20           & 72.80           & 47.70            & 88.30          & 70.30          & 1503           & 69.90          \\
SAD-ENet~\cite{SAD}                                 & 70.80          & 33             & 90.10           & 68.80            & 60.20           & 65.90           & 41.60            & 84.00          & 65.70          & 1998           & 66.00          \\
LaneAF-DLA34~\cite{LaneAF}                             & 77.41          & 28             & 91.80           & 75.61            & 71.78           & \textbf{79.12}  & \textbf{51.38}   & 86.88          & \textbf{72.70} & 1360           & 73.03          \\
AtrousFormer-R34~\cite{AtrousFormer}                       & \textbf{78.08} & -             & \textbf{92.83}  & \textbf{75.96}   & \textbf{69.48}  & 77.86           & 50.15            & \textbf{88.66} & 71.14          & \textbf{1054}  & \textbf{73.74} \\ \cline{1-1}
\textbf{Keypoint Based}                  &                &              &                 &                  &                 &                 &                  &                &                &                &                \\ \cline{1-1}
PINet-Hourglass~\cite{PINet}                          & 74.40          & 27              & 90.30           & 72.30            & 66.30           & 68.40           & 49.80            & 83.70          & 65.20          & 1427           & 67.70          \\
FOLOLane-ERFNet~\cite{FOLOLane}                          & 78.80          & -             & 92.70           & 77.80            & \textbf{75.20}  & 79.30           & 52.10            & 89.00          & 69.40          & 1569           & \textbf{74.50} \\
GANet-R34~\cite{GANnet}                              & \textbf{79.39} & 69             & \textbf{93.73}  & \textbf{77.92}   & 71.64           & \textbf{79.49}  & \textbf{52.63}   & \textbf{90.37} & \textbf{76.32} & \textbf{1368}  & 73.67          \\ \cline{1-1}
\textbf{Parameter Based}                 &                &              &                 &                  &                 &                 &                  &                &                &                &                \\ \cline{1-1}
BézierLaneNet-R34~\cite{BezierLaneNet}                      & 75.57          & 78             & 91.59           & 73.20            & 69.20           & \textbf{76.74}  & 48.05            & 87.16          & 62.45          & 888            & 69.90          \\
Laneformer-R50~\cite{Laneformer}                         & \textbf{77.06} & -              & \textbf{91.77}  & \textbf{75.41}   & \textbf{70.17}  & 75.75           & \textbf{48.73}   & \textbf{87.65} & \textbf{66.33} & \textbf{19}    & \textbf{71.04} \\ \cline{1-1}
\textbf{Anchor \& Detection Based} &                &              &                 &                  &                 &                 &                  &                &                &                &                \\ \cline{1-1}
FastDraw-R50~\cite{fastdraw}                           & -              & -             & 85.90           & 63.60            & 57.00           & 69.90           & 40.60            & 79.40          & 65.20          & 7013           & 57.80          \\
UFLDv2-R34~\cite{UFLDv2}                             & 76.00          & \textbf{114}             & 92.50           & 74.80            & 65.50           & 75.50           & 49.20            & 88.80          & 70.10          & 1910           & 70.80          \\
CurveLanes-L~\cite{curvelanes}                             & 74.80          & -             & 90.70           & 72.30            & 67.70           & 70.10           & 49.40            & 85.80          & 68.40          & 1746           & 68.90          \\
LaneATT-R122~\cite{Laneatt}                           & 77.02          & 38           & 91.74           & 76.16            & 69.47           & 76.31           & 50.46            & 86.29          & 64.05          & 1264           & 70.81          \\
SGNet-R34~\cite{SGNet}                              & 77.27          & -             & 92.07           & 75.41            & 67.75           & 74.31           & 50.90            & 87.97          & 69.65          & 1373           & 72.69          \\
CondLane-R34~\cite{ConditionLane}                           & 78.74          & 70          & {\ul 93.38}     & 77.14            & 71.17           & \textbf{79.93}  & 51.85            & 89.89          & \textbf{73.88} & 1387           & 73.92          \\
CLRNet-R34~\cite{Clrnet}                             & \textbf{79.73} & 63           & \textbf{93.49}  & \textbf{78.06}   & \textbf{74.57}  & {\ul 79.92}     & \textbf{54.01}   & \textbf{90.59} & {\ul 72.77}    & {\ul 1216}     & \textbf{75.02} \\ \hline
\rowcolor{mygray} 
\textbf{ADNet-R18 (Ours)}             & 77.56          & {\ul 87}          & 91.92           & 75.81            & 69.39           & 76.21           & 51.75            & 87.71          & 68.84          & \textbf{1133}  & 72.33          \\
\rowcolor{mygray}
\textbf{ADNet-R34 (Ours)}             & {\ul 78.94}    & 77           & 92.90           & {\ul 77.45}      & {\ul 71.71}     & 79.11           & {\ul 52.89}      & {\ul 89.90}    & 70.64          & 1499           & {\ul 74.78}    \\ \hline
\end{tabular}
}
\end{center}
\vspace{-6mm}
\end{table*}
\subsection{Performance on benchmarks}

\textbf{VIL-100.} Our approach achieves state-of-the-art results on the recently released VIL-100 lane detection dataset. In Table \ref{tb.vil}, we compare our results with the previous state-of-the-art method MMA-Net~\cite{VIL-100}, and show that our method has increased F1@50 from 83.90 to 90.90. We have also achieved a lane accuracy of 94.27 with ResNet101 and 94.38 with ResNet34, which is much better than MMA-Net. Our results have also compared with the anchor-free state-of-the-art method CLRNet, which performs very well on multiple benchmarks such as CULane and LLAMAS~\cite{LLamas}. However, on VIL-100, CLRNet fails to maintain its edge. To provide a fair comparison with CLRNet, we have relocated our start points into three edges of the image, and our smallest model has outperformed CLRNet in multiple indicators.
\begin{figure}[t]
\begin{center}
\includegraphics[width=1.\linewidth]{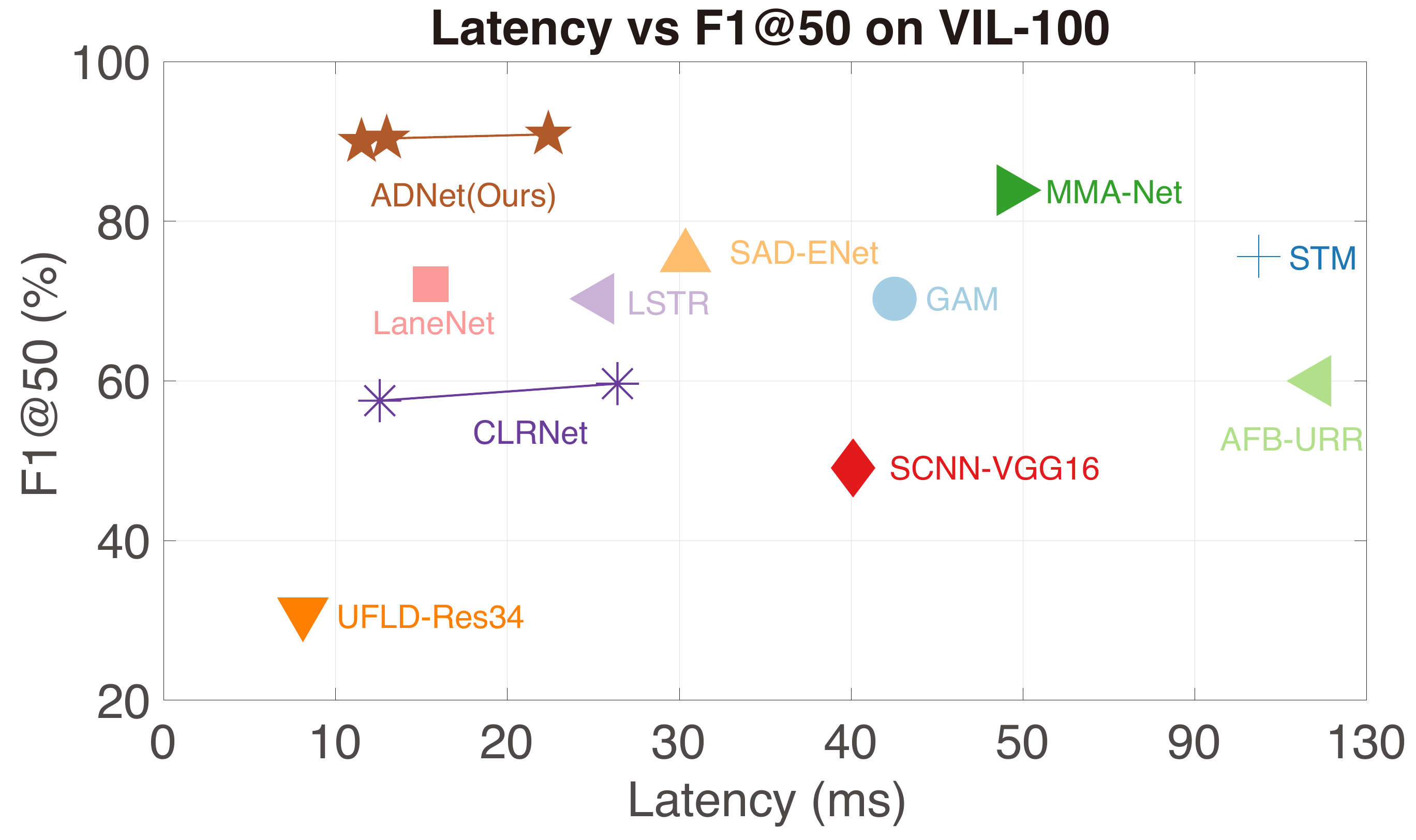}
\end{center}
\vspace{-5mm}
   \caption{Latency vs F1@50 of other methods on VIL-100 lane detection benchmark. Our method outperforms all existing methods and maintains a promising inference speed.}
\label{speed-acc}
\vspace{-5mm}
\end{figure}

\begin{table}[h!]
\begin{center}
\caption{Comparison with state-of-art methods on VIL-100 test set. 
{Our proposed ADNet are flexible in modelling the locations of start points (they can be anywhere in the images). For more comparisons, we also provide a ``ADNet$^*$'' version where the start points are extended to the three edges likewise in CLRNet, showing inferior performance.}
}
\vspace{-2mm}
\label{tb.vil}
\resizebox{\linewidth}{!}
{
\begin{tabular}{cccccc}
\hline
\textbf{Methods}                     & \textbf{F1@50} $\uparrow$ & \textbf{Acc} $\uparrow$ & \textbf{FP} $\downarrow$ & \textbf{FN} $\downarrow$ & \textbf{FPS} $\uparrow$ \\ \hline
\textbf{VOS Methods}                 &                     &                        &                  &                  &              \\ \cline{1-1}
GAM~\cite{GAM}                                  & 70.30               & 85.50                  & 24.1             & 21.2             & 24             \\
RVOS~\cite{RVOS}                                 & 51.90               & 90.90                  & 61.0             & 11.9             & -             \\
STM~\cite{STM}                                  & 75.60               & 90.20                  & 22.8             & 12.9             & 10            \\
AFB-URR~\cite{AFB-URR}                              & 60.00               & 84.60                  & 25.5             & 22.2             & 9             \\
TVOS~\cite{TVOS}                                 & 24.00               & 46.10                  & 58.2             & 62.1             & 36              \\
MMA-Net~\cite{VIL-100}                              & \textbf{83.90}      & \textbf{91.00}         & \textbf{11.1}    & \textbf{10.5}    & 20             \\ \cline{1-1}
\textbf{\makecell[c]{Lane Detection \\ Methods}}      &                     &                        &                  &                  &              \\ \cline{1-1}
LaneNet~\cite{LaneNet}                              & 72.10               & 85.80                  & 12.2             & 20.7             & 64             \\
SCNN-VGG16~\cite{SCNN}                                 & 49.10               & 90.70                  & 12.8             & 11.0             & 25             \\
SAD-ENet~\cite{SAD}                            & 75.50               & 88.60                  & 17.0             & 15.2             & 33             \\
UFLD-R34~\cite{UFLD}                                 & 31.00               & 85.20                  & 11.5             & 21.5             & \textbf{124}             \\
LSTR~\cite{LSTR}                                 & 70.30               & 88.40                  & 16.3             & 14.8             & 40             \\
CLRNet-R18~\cite{Clrnet}                         & 57.27               & 88.99                  & 6.9              & 13.5             & 80             \\
CLRNet-R101~\cite{Clrnet}                        & 59.41               & 88.65                  & \textbf{2.1}     & 12.5             & 38             \\ \hline
ADNet-R18$^*$ (Ours) & 65.05               & 94.25                  & 5.0              & 5.0              & -             \\
ADNet-R34$^*$ (Ours) & 64.97               & {\ul 94.37}                  & 4.5              & 4.9              & -             \\
\rowcolor{mygray}
\textbf{ADNet-R18 (Ours)}         & 89.97               & 94.23                  & 5.0              & 5.1              & {\ul 87}             \\

\rowcolor{mygray}
\textbf{ADNet-R34 (Ours)}         & {\ul 90.39}               & \textbf{94.38}         & {\ul 4.4}        & \textbf{4.9}     & 77             \\

\rowcolor{mygray}
\textbf{ADNet-R101 (Ours)}        & \textbf{90.90}      & 94.27                  & 4.7             & {\ul 5.0}             & 45             \\ \hline
\end{tabular}
}
\end{center}
\vspace{-6mm}
\end{table}

\textbf{CULane.} The performance of ADNet is compared with other state-of-the-art methods on CULane and the results are presented in Table \ref{tb.culane}. 
Compared to the previous fixed anchor method, for example, LaneATT~\cite{Laneatt}, our method achieves a convincing F1 score of 78.94 with ResNet34, outperforming LaneATT with ResNet122 by 1.92\%. Our approach also surpasses subsequent anchor-free techniques, such as SGNet~\cite{SGNet}, by 1.67\%. Additionally, our method achieves state-of-the-art performance among segmentation-based and parameter-based methods, but is ranked second after CLRNet~\cite{Clrnet} on Anchor \& Detection-based methods.

\textbf{TuSimple.} The comparison results on TuSimple are presented in Table \ref{tb.tusimple}. Due to the limited scenario of TuSimple, the differences between each method are minimal. As can be observed from the table, our method performs better than most of the compared methods.

\begin{table}[h!]
\begin{center}
\caption{Comparison with state-of-art methods on TuSimple test set.}
\vspace{-2mm}
\label{tb.tusimple}
\resizebox{\linewidth}{!}
{
\begin{tabular}{ccccc}
\hline
\textbf{Methods}             & \textbf{F1@50} $\uparrow$ & \textbf{Acc} $\uparrow$ & \textbf{FP} $\downarrow$ & \textbf{FN} $\downarrow$ \\ \hline
SCNN~\cite{SCNN}                         & 95.97               & 96.53                  & 6.17             & \textbf{1.80}    \\
RESA-R50~\cite{RESA}                   & 96.93               & 96.82                  & 3.63             & 2.48             \\
PolyLaneNet~\cite{PolyLaneNet}                  & 90.62               & 93.36                  & 9.42             & 9.33             \\
E2E-ERFNet~\cite{e2e}                   & 96.25               & 96.02                  & 3.21             & 4.28             \\
UFLD-R34~\cite{UFLD}                   & 88.02               & 95.86                  & 18.91            & 3.75             \\
UFLDv2-R34~\cite{UFLDv2}                 & 96.22               & 95.56                  & 3.18             & 4.37             \\
SGNet-R34~\cite{SGNet}                  & -                   & 95.87                  & -                & -                \\
LaneATT-R34~\cite{Laneatt}                & 96.77               & 95.63                  & 3.53             & 2.92             \\
CondLaneNet-R101~\cite{ConditionLane}           & 97.24               & 96.54                  & \textbf{2.01}    & 3.50             \\
FOLOLane-ERFNet~\cite{FOLOLane}              & 96.59               & \textbf{96.92}         & 4.47             & 2.28             \\ \hline
\rowcolor{mygray}
\textbf{ADNet-R18 (Ours)} & 96.90               & 96.23                  & 2.91             & 3.29             \\
\rowcolor{mygray}
\textbf{ADNet-R34 (Ours)} & \textbf{97.31}      & 96.60                  & 2.83             & 2.53             \\ \hline
\end{tabular}
}
\end{center}
\vspace{-6mm}
\end{table}

\subsection{Ablation studies}

\textbf{Overall.} We conduct overall ablation study using ResNet18 as the backbone in CULane. The baseline model extracts features from the backbone and FPN, pooling lane features according to the strategy in~\cite{Mask-rcnn}, identical to ADNet, and regressing lane lines using LIoU loss with predefined anchors from~\cite{LineCNN}. We gradually add SPGU, ALAU, and GLIoU to the baseline, and finally embed LKA with FPN. The overall ablation study results in Table \ref{tb.overall_ab} show that GLIoU has the least effect, while SPGU has the greatest effect. The remaining strategies has effects that ranged from big to small, namely ALAU and LKA. Adding SPGU significantly increases the F1@50 score from 72.17 to 76.47, strongly supporting our assumption.
\begin{table}[h!]
\begin{center}
\caption{Overall ablation study of ADNet-R18 on CULane.}
\vspace{-2mm}
\label{tb.overall_ab}
\resizebox{\linewidth}{!}
{
\begin{tabular}{cccccc}
\hline
\textbf{Baseline} & \textbf{+SPGU} & \textbf{+ALAU} & \textbf{+GLIoU} & \textbf{+LKA} & \textbf{F1@50}                 \\ \hline
$\surd$        &       &       &        &       &72.17                        \\
         & $\surd$       &       &        &       & 76.47 (+4.30)          \\
         & $\surd$       & $\surd$      &        &       & 76.91 (+4.74)          \\
         & $\surd$      & $\surd$      & $\surd$        &       & 77.15 (+4.98)          \\
         & $\surd$       & $\surd$       & $\surd$       & $\surd$       & \textbf{77.56 (+5.39)} \\ \hline
\end{tabular}
}
\end{center}
\vspace{-8mm}
\end{table}

\begin{figure*}
    \centering
    \includegraphics[width=\linewidth]{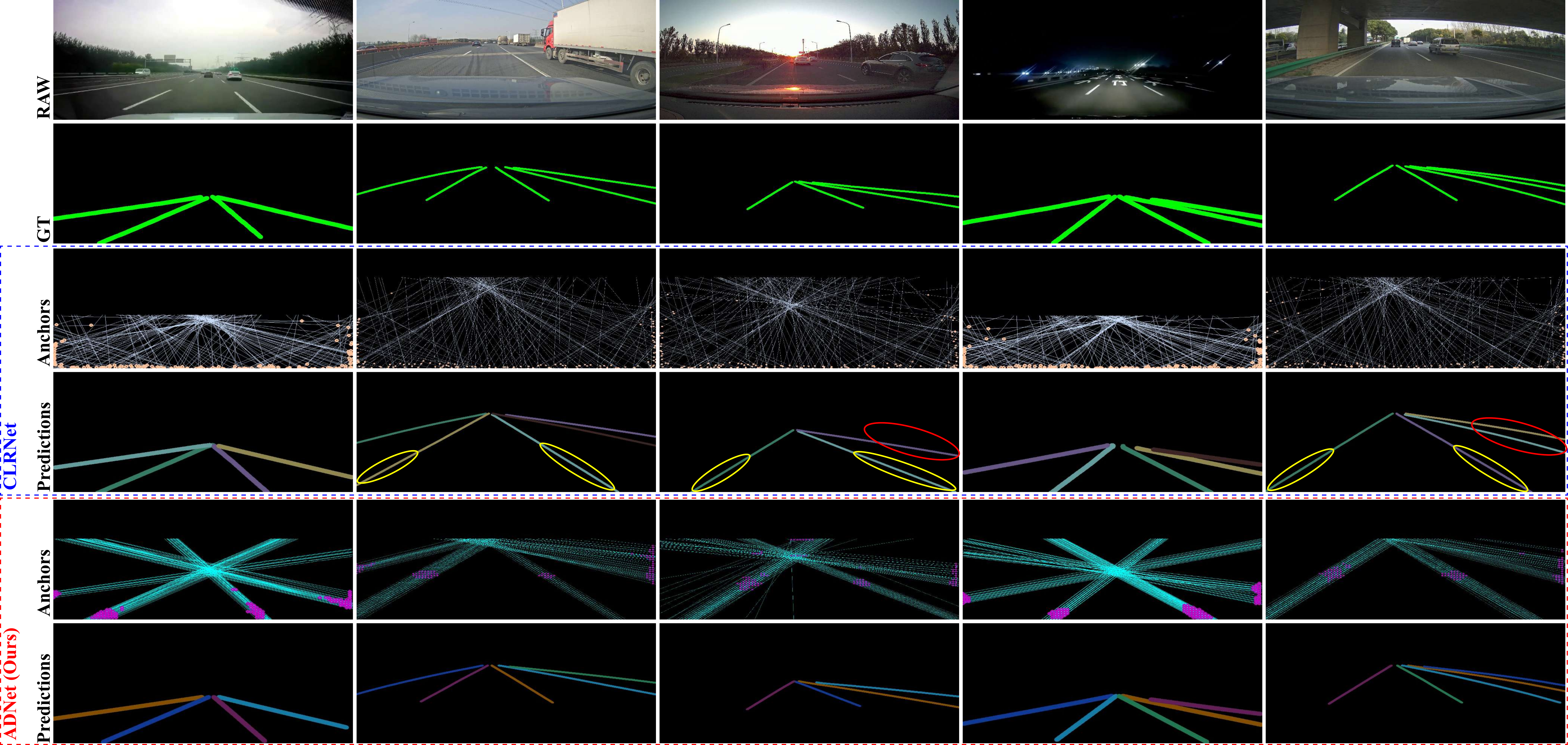}
    \vspace{-5mm}
    \caption{Visualisation results on VIL-100 compare with CLRNet. We visualised every anchors before predictions, \textcolor{yellow}{yellow} oval is applied to highlight the anchor flexibility issue discussed on Section~\ref{chp.introduction}. It is evident that our anchors exhibit higher quality compared to CLRNet, which consequently leads to better performance, highlighted by \textcolor{red}{red} oval.}
    \vspace{-5mm}
\end{figure*}

\textbf{Effectiveness of guidance map.} In ALAU, kernel offsets are obtained from the guidance map as discussed in Section~\ref{chp.alau}. This allows us to promote $f_{v}: \Theta_{xy} \to \vec{v}$ to $f_{v}: (\Theta_{xy},P_{xy}) \to \vec{v}$, as the starting point and its theta value are spatially related. Our experiments on CULane and VIL-100, as shown in Table \ref{tb.ablation_parts}, confirm this conclusion. Without guidance map represents that we obtain kernel offsets simply from thetas map. The result indicates that when guidance map is added, improvement can be observed on both benchmarks.

\textbf{Necessity of GLIoU loss.} In our overall ablation study, only a 0.24\% improvement is brought by GLIoU loss compared to LIoU loss, which is explainable. The scenario we describe in Section~\ref{chp.gliou} rarely occurs since on CULane lane lines always ray from three edges of the image. Further experiments on CULane (shown in Table \ref{tb.ablation_parts}) demonstrate that switching the backbone from ResNet18 to ResNet34 with GLIoU loss only brings a 0.18\% increment, similar to the phenomenon in Table \ref{tb.overall_ab}. However, when we conduct the same experiments on VIL-100, both ResNet18 and ResNet34 get a huge boost on F1@50.

\begin{table}[h!]
\begin{center}
\caption{Ablation study on different components. ``w/o'' under \textbf{Guidance map} represents obtaining kernel offsets from thetas map; ``baseline'' under \textbf{Attention} follows~\cite{conv2former}. }
\vspace{-2mm}
\label{tb.ablation_parts}
\resizebox{\linewidth}{!}
{
\begin{tabular}{cccccc}
\hline
\textbf{Dataset}          & \textbf{\makecell[c]{Back- \\ bone}} & \textbf{\makecell[c]{Guidance \\ map}}     & \textbf{Loss}                 & \textbf{Attention}          & \textbf{F1@50}                                                          \\ \hline
                          & R34          & \cellcolor[HTML]{FFF2CC}w/o & GLIoU                         & baseline                         & \cellcolor[HTML]{FFF2CC}78.53                                           \\
                          & R34          & \cellcolor[HTML]{FFE699}w   & GLIoU                         & baseline                         & \cellcolor[HTML]{FFE699}{\color[HTML]{333333} \textbf{78.66   (\textcolor{red}{+0.13})}} \\ \cline{2-6} 
                          & R34          & w                           & \cellcolor[HTML]{DDEBF7}LIoU  & baseline                         & \cellcolor[HTML]{DDEBF7}78.48                                           \\
                          & R34          & w                           & \cellcolor[HTML]{BDD7EE}GLIoU & baseline                         & \cellcolor[HTML]{BDD7EE}\textbf{78.66   (\textcolor{red}{+0.18})}                        \\ \cline{2-6} 
                          & R34          & w                           & GLIoU                         & \cellcolor[HTML]{E2EFDA}baseline & \cellcolor[HTML]{E2EFDA}78.66                                           \\
\multirow{-6}{*}{CULane}  & R34          & w                           & GLIoU                         & \cellcolor[HTML]{C6E0B4}LKA     & \cellcolor[HTML]{C6E0B4}\textbf{78.94   (\textcolor{red}{+0.28})}                        \\ \hline
                          & R34          & \cellcolor[HTML]{FFF2CC}w/o & GLIoU                         & baseline                         & \cellcolor[HTML]{FFF2CC}87.83                                           \\
                          & R34          & \cellcolor[HTML]{FFE699}w   & GLIoU                         & baseline                         & \cellcolor[HTML]{FFE699}\textbf{89.22   (\textcolor{red}{+1.39})}                        \\ \cline{2-6} 
                          & R34          & w                           & \cellcolor[HTML]{DDEBF7}LIoU  & LKA                             & \cellcolor[HTML]{DDEBF7}90.17                                           \\
                          & R34          & w                           & \cellcolor[HTML]{BDD7EE}GLIoU & LKA                             & \cellcolor[HTML]{BDD7EE}\textbf{90.39   (\textcolor{red}{+0.22})}                       \\
                          & R18          & w                           & \cellcolor[HTML]{DDEBF7}LIoU  & baseline                         & \cellcolor[HTML]{DDEBF7}83.44                                           \\
                          & R18          & w                           & \cellcolor[HTML]{BDD7EE}GLIoU & baseline                         & \cellcolor[HTML]{BDD7EE}\textbf{88.65   (\textcolor{red}{+5.21})}                        \\ \cline{2-6} 
                          & R34          & w                           & GLIoU                         & \cellcolor[HTML]{E2EFDA}baseline & \cellcolor[HTML]{E2EFDA}89.22                                           \\
\multirow{-8}{*}{VIL-100} & R34          & w                           & GLIoU                         & \cellcolor[HTML]{C6E0B4}LKA     & \cellcolor[HTML]{C6E0B4}\textbf{90.39   (\textcolor{red}{+1.17})}                        \\ \hline
\end{tabular}
}
\end{center}
\vspace{-9mm}
\end{table}

\textbf{Ablation study of LKA.} We validate the effectiveness of our LKA by conducting experiments with different backbones and datasets. Results in Table \ref{tb.ablation_parts} and Table \ref{tb.overall_ab} indicate that our LKA not only improves upon plain FPN, but also outperforms the baseline attention module on both VIL-100 and CULane.


\vspace{-3mm}
\section{Conclusion}
\vspace{-2mm}
In this paper, we propose ADNet for lane shape prediction, incorporating SPGU to predict start points and ALAU to aggregate context near lane lines. We introduce GLIoU loss to address limitations of LIoU loss and modify the small kernel attention module into LKA. Our algorithm outperforms current state-of-the-art methods on VIL-100 and achieves nearest state-of-the-art on CULane and TuSimple.

\section*{Acknowledgement}
This work was supported by the National Natural Science Foundation of China under No.62276061 and 62006041.

{\small
\bibliographystyle{ieee_fullname}
\bibliography{lane}
}

\appendix
\newpage

\maketitle
\ificcvfinal\thispagestyle{empty}\fi

\section{Representation of lane and lane anchor}
On an image, a lane can be represented by the x-coordinates of 2D points that are equidistantly sliced by the y-axis~\cite{LineCNN}. We denote a lane as $l = \{x_1, x_2, ..., x_k\}$, where $x_i$ represents the x-coordinate of the $i$-th slice and $k$ is the total number of slices. Since the slicing scheme is fixed and equidistant (from bottom to top), the y-coordinates are distributed on a fixed pattern. Therefore, a set of y-coordinates for every lane on the image can be generated as $Y = \{y_1, y_2, ..., y_k\}$. With $l$ and $Y$, we can locate the positions of points that construct a lane.

A lane anchor can be noted as $(s_x, s_y, \theta)$, where $s_x$ and $s_y$ represent the x-coordinates and y-coordinates of the start point, respectively, and $\theta$ represents the angle of the start point. With $(s_x,s_y,\theta)$, we can describe lane anchor as a ray using:
\begin{equation}
    x_i = s_x \cdot w + \frac{(y_i-s_y)h}{tan(1-\theta)\cdot \pi}\ , \ y_i \in Y,
\end{equation}
where $w$ and $h$ are the width and height of the image, $\theta$ is the angle between the anchor and the x-axis in a clockwise direction.

\begin{table}[h!]
\begin{center}
\vspace{-1mm}
\caption{Effectiveness of different MSA designs. Models are trained and tested on CULane~\cite{SCNN} using the ResNet34 as backbone.}
\vspace{-3mm}
\label{tb.lksa_ab}
\resizebox{\linewidth}{!}
{
\begin{tabular}{lc}
\hline
\multicolumn{1}{c}{\textbf{MSA Design}}                                                                     & \textbf{F1@50(\%)} \\ \hline
without any attention module    &78.52\\
baseline                                                                                                     & 78.66 \\
\makecell[l]{MSA-A: Replace $DConv_{11\times11}$ \\ \quad with multi-channel strip $DConv$}                                          & 78.46 \\
MSA-B: Add indentical forward path                                                                                & 78.34 \\
\rowcolor{mygray}
MSA-C: Replace Liner with $DConv_{5\times5}$ & \textbf{78.94} \\ \hline
\end{tabular}
}
\end{center}
\vspace{-8mm}
\end{table}

\section{Micro design of LKA}\label{chp.App_LKA}
\begin{figure*}

\subfigure[baseline]{
\begin{minipage}[t]{0.13\linewidth}
\centering
\includegraphics[width=0.7in]{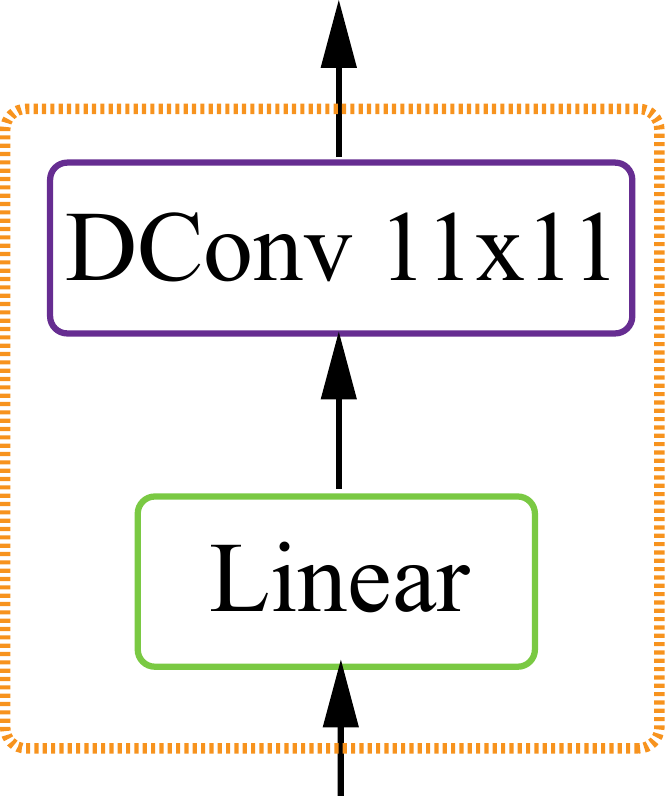}
\label{MSA-baseline}
\end{minipage}%
}%
\subfigure[MSA-A]{
\begin{minipage}[t]{0.29\linewidth}
\centering
\includegraphics[width=1.8in]{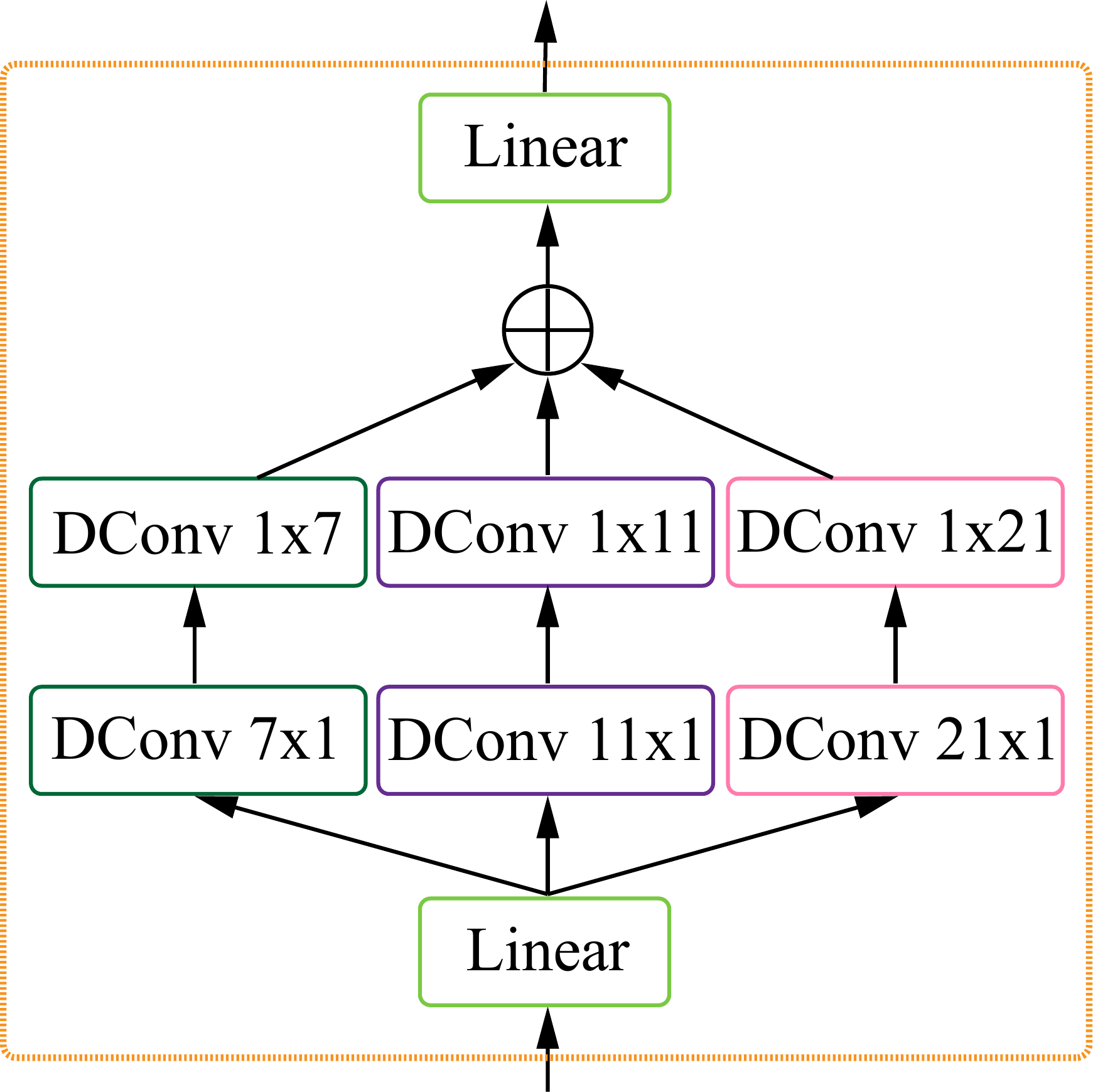}
\label{MSA-a}
\end{minipage}%
}%
\subfigure[MSA-B]{
\begin{minipage}[t]{0.29\linewidth}
\centering
\includegraphics[width=1.9in]{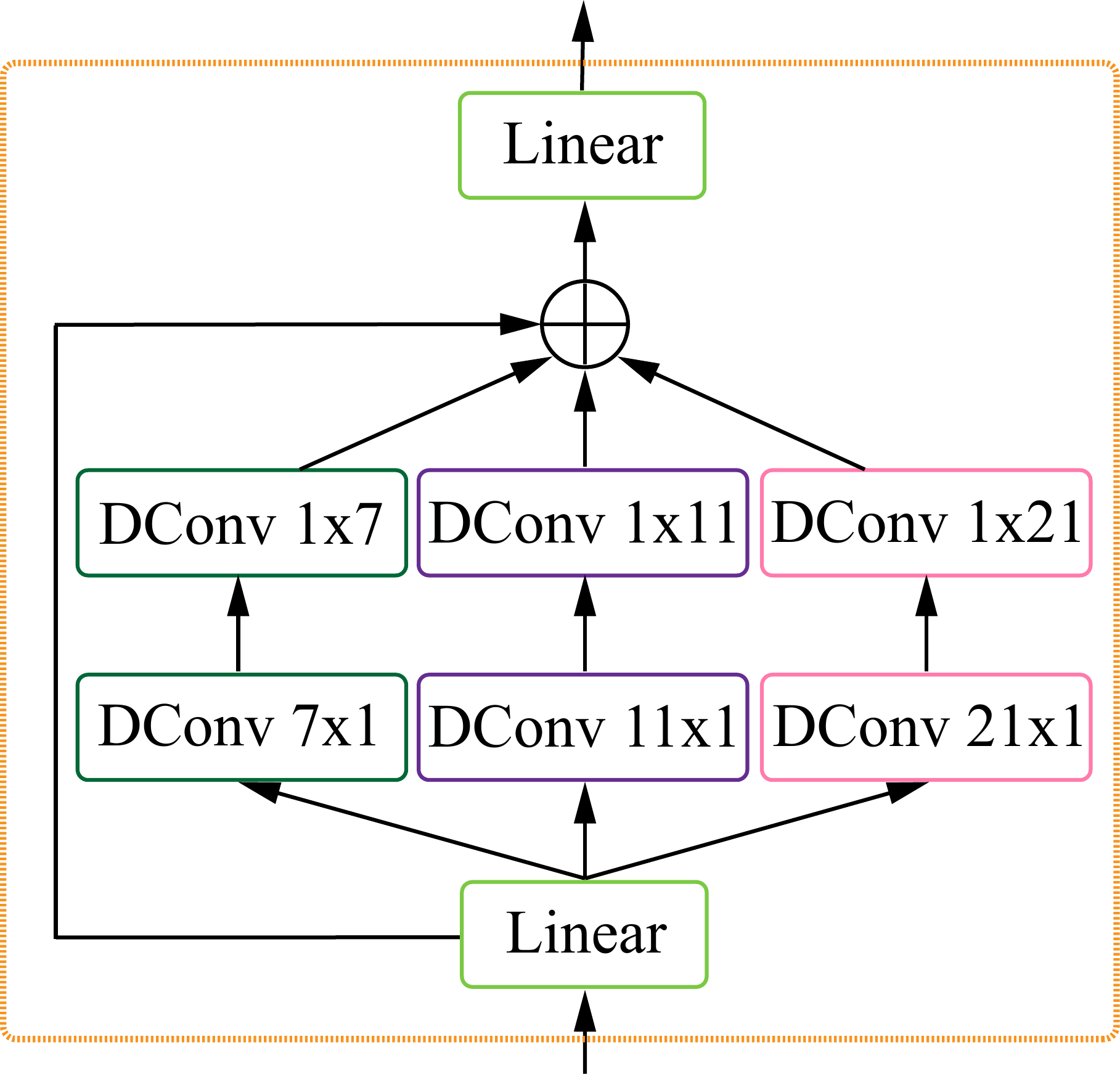}
\label{MSA-b}
\end{minipage}%
}%
\subfigure[MSA-C]{
\begin{minipage}[t]{0.29\linewidth}
\centering
\includegraphics[width=1.9in]{figs/SA_MSA.pdf}
\label{MSA-c}
\end{minipage}%
}%
\vspace{-2mm}
\caption{Illustration of different designs of MSA.}
\label{MSA_design}
\vspace{-5mm}
\end{figure*}

To best take advantage of a large kernel, we mainly focus on discovering an effective way to generate the attention matrix $Att$ for the lane detection task. We demonstrate our design by gradually reconstructing the baseline (originating from~\cite{conv2former}) to our final LKA. Apart from the baseline, Figure~\ref{MSA_design} presents our three candidate designs of the MSA module on LKA. MSA-A (Figure~\ref{MSA-a}) replaces the $DConv$ on the baseline with a multi-channel strip $DConv$, MSA-B (Figure~\ref{MSA-b}) further adds an identical forward path, and MSA-C (Figure~\ref{MSA-c}) replaces Liner with $DConv$ with a kernel size of $5\times5$. A series of experiments are shown in Table~\ref{tb.lksa_ab}. Our final mechanism, MSA-C, exhibits superior performance to the other designs.

\begin{figure}[t]
\vspace{-3mm}
\includegraphics[width=\linewidth]{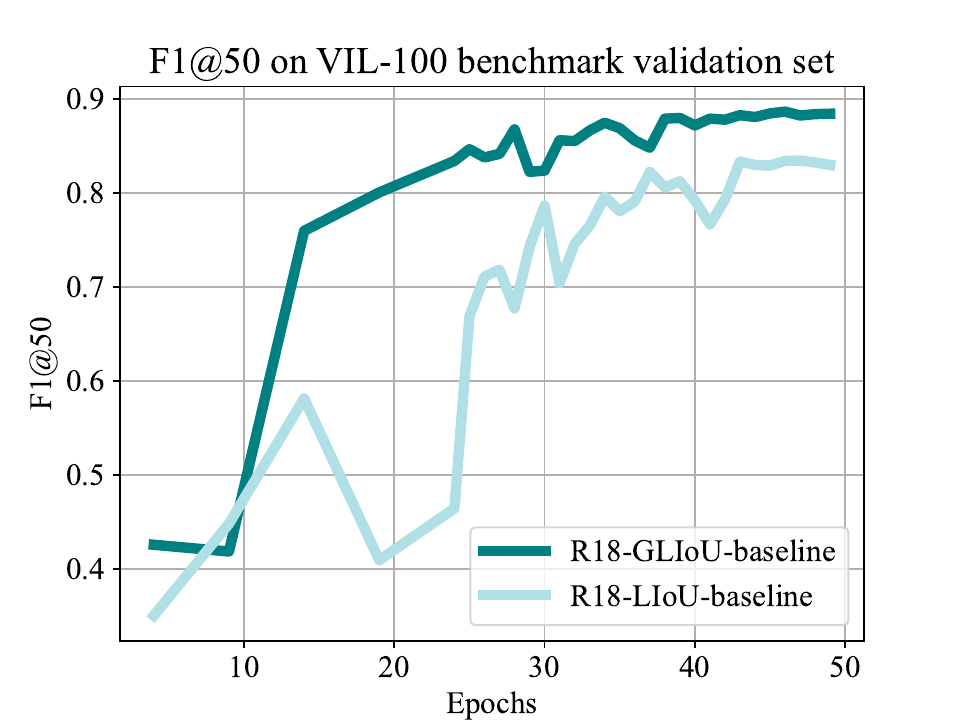}
\vspace{-3mm}
\caption{Illustration of the effectiveness of GLIoU on VIL-100.}
\label{GLIoU-LIoU}
\vspace{-5mm}
\end{figure}
\section{GLIoU}
To emphasise the effectiveness of GLIoU, we present the results on the validation set of VIL-100~\cite{VIL-100} in chart form. 
The validation results are presented in Figure~\ref{GLIoU-LIoU}. ``R18-GLIoU-baseline'' represents ResNet18~\cite{Resnet} as the backbone, GLIoU as the regression loss, and ``baseline''~\cite{conv2former} as the attention module.

The results in Figure~\ref{GLIoU-LIoU} demonstrate that using GLIoU as the regression loss function generally yields better performance than LIoU~\cite{Clrnet} across all documented epochs and reaches saturation more quickly than LIoU.

\begin{table}[h!]
\begin{center}
\caption{Effectiveness of anchors number. Models are trained and tested on CULane and VIL-100 using different backbones. ``R18'' stands for ResNet18.}
\resizebox{\linewidth}{!}
{
\begin{tabular}{cccccc}
\hline
\textbf{Dataset}         & \textbf{\begin{tabular}[c]{@{}c@{}}Back-\\ bone\end{tabular}} & \makecell[c]{\textbf{Anc-} \\ \textbf{hors}} & \textbf{F1@50(\%)} & \textbf{Precision(\%)} & \textbf{Acc(\%)} \\ \hline
\multirow{5}{*}{CULane}  & \multirow{5}{*}{R18}                                          & 100              & 77.53          & 85.56              & -            \\
                         &                                                               & 200              & 77.51          & 85.20              & -            \\
                         &                                                               & 300              & 77.56          & 85.01              & -            \\
                         &                                                               & 400              & 77.63          & 84.85              & -            \\
                         &                                                               & 500              & 77.61          & 86.62              & -            \\ \hline
\multirow{6}{*}{VIL-100} & \multirow{3}{*}{R18}                                          & 50               & 90.02          & 90.50              & 94.10        \\
                         &                                                               & 75               & 90.04          & 90.29              & 94.21        \\
                         &                                                               & 100              & 89.97          & 90.09              & 94.23        \\ \cline{2-6} 
                         & \multirow{3}{*}{R101}                                         & 50               & 90.83          & 91.10              & 94.14        \\
                         &                                                               & 75               & 90.90          & 90.99              & 94.27        \\
                         &                                                               & 100              & 90.90          & 90.99              & 94.27        \\ \hline
\end{tabular}
}
\label{tb.ank_num}
\end{center}
\end{table}

\section{Supervision for heat map}
\begin{figure}[t]

\subfigure[$\sigma=1$]{
\begin{minipage}[t]{0.5\linewidth}
\centering
\includegraphics[width=1.6in]{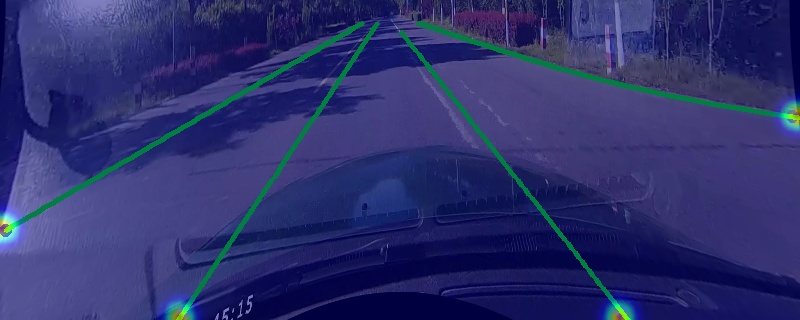}
\end{minipage}%
}%
\subfigure[$\sigma=2$]{
\begin{minipage}[t]{0.5\linewidth}
\centering
\includegraphics[width=1.6in]{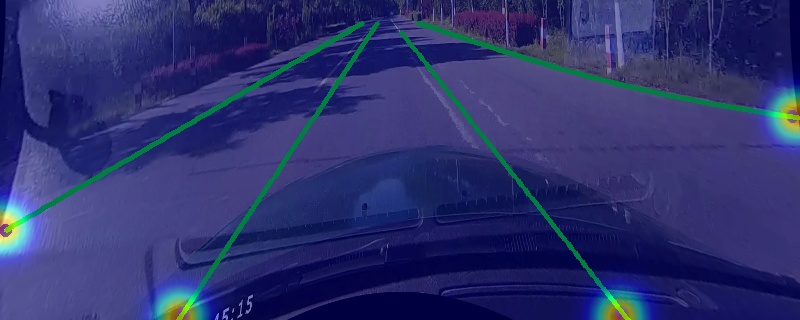}
\end{minipage}%
}%
\vspace{-2mm}
\quad
\subfigure[$\sigma=4$]{
\begin{minipage}[t]{0.5\linewidth}
\centering
\includegraphics[width=1.6in]{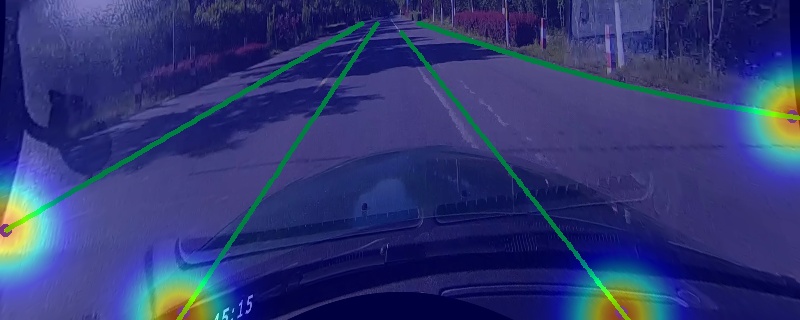}
\end{minipage}%
}%
\subfigure[$\sigma=8$]{
\begin{minipage}[t]{0.5\linewidth}
\centering
\includegraphics[width=1.6in]{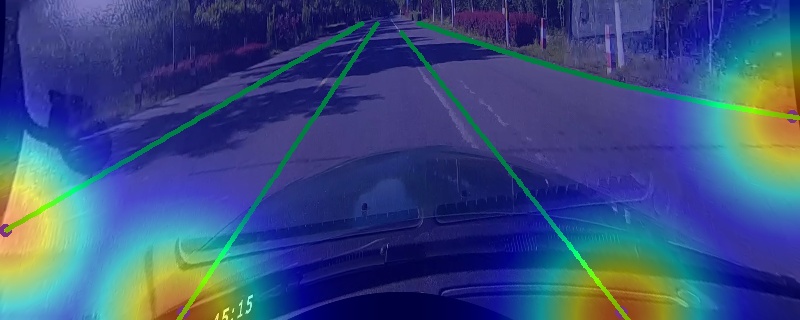}
\end{minipage}%
}%
\vspace{-3mm}
\caption{Illustration of start points heat map supervision with different $\sigma$ on CULane.}
\vspace{-5mm}
\label{Supervision}
\end{figure}

In our framework, we propose supervision for the start points heat map using a non-normalised Gaussian kernel. The hyperparameter $\sigma$ controls the size of the region that potentially contains start points. Scenarios that typically consist of invisible start points tend to favour higher $\sigma$ values compared to scenarios with clear and eye-catching start points. Figure~\ref{Supervision} visualises the start points heat map supervision using four different $\sigma$ values on CULane. On the activation map, red regions potentially contain start points, while blue regions have a lower likelihood.

Table~\ref{tb.sigma} shows the results of our experiments conducted on CULane and VIL-100. For CULane, we choose $\sigma$ values of 2, 4, and 8 since start points on CULane are usually invisible. For VIL-100, we choose $\sigma$ values of 1, 2, and 4 since start points are generally clear and featured in this dataset. It can be deduced that when $\sigma$ on CULane is larger than 4, limited improvement can be observed, while $\sigma$ lower than 4 causes a reduction in performance. VIL-100 presents an opposite tendency.

\begin{table}[h!]
\begin{center}
\vspace{-2mm}
\caption{Effectiveness of different hyperparameter $\sigma$ for start points heat map supervision. Models are trained and tested on CULane and VIL-100 using the ResNet18 as backbone.}
\label{tb.sigma}
\resizebox{\linewidth}{!}
{
\begin{tabular}{ccccccc}
\hline
\textbf{Dataset}                  & $\mathbf{\sigma}$ &\textbf{F1@50(\%)}   & \textbf{Precision(\%)} & \textbf{Acc(\%)}  \\ \hline
\multirow{3}{*}{CULane}  & 2        & 76.87          & 83.60     & -     \\
                         & 4        & \textbf{77.56} & 85.01     & -       \\
                         & 8        & 77.51          & \textbf{86.25}     & -     \\ \hline
\multirow{3}{*}{VIL-100} & 1        & 89.69          & \textbf{90.24}     & 94.14    \\
                         & 2        & \textbf{89.97} & 90.09     & \textbf{94.23}    \\
                         & 4        & 86.66          & 85.94     & 94.04   \\ \hline
\end{tabular}
}
\end{center}
\vspace{-5mm}
\end{table}

\section{Number of anchors}

The number of anchors can be regarded as a hyperparameter during the training and validation phases. Our experiments are conducted on CULane and VIL-100, using different backbones, ResNet18 and ResNet101. The training phase follows the procedure outlined in our paper. Results are presented in Table~\ref{tb.ank_num}. Increasing the number of anchors does not necessarily lead to improved performance; in fact, it can even lead to a decrease. Therefore, to strike a balance across multiple indicators, the number of anchors can vary.

\begin{figure*}

\subfigure[Visualisation results on VIL-100]{
    \centering
    \includegraphics[width=\linewidth]{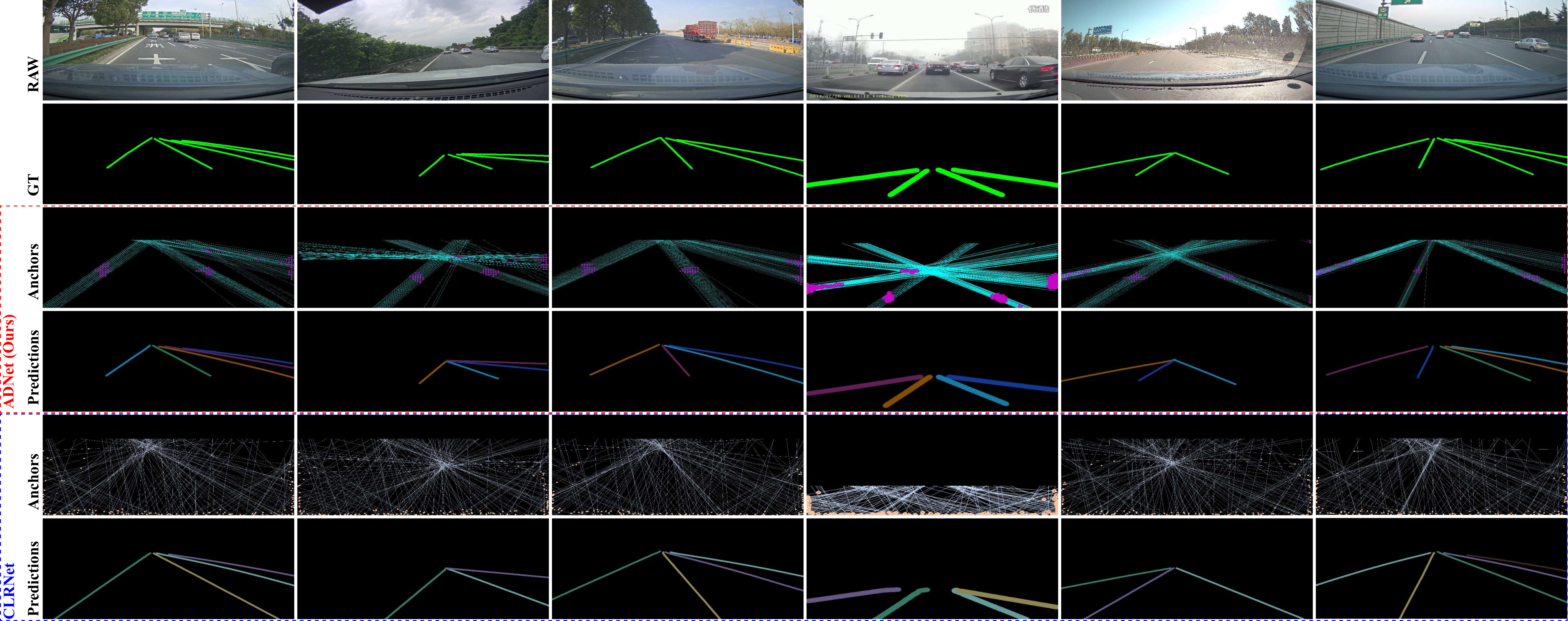}
}
\vspace{-3mm}
\quad

\subfigure[Visualisation results on CULane]{
    \centering
    \includegraphics[width=\linewidth]{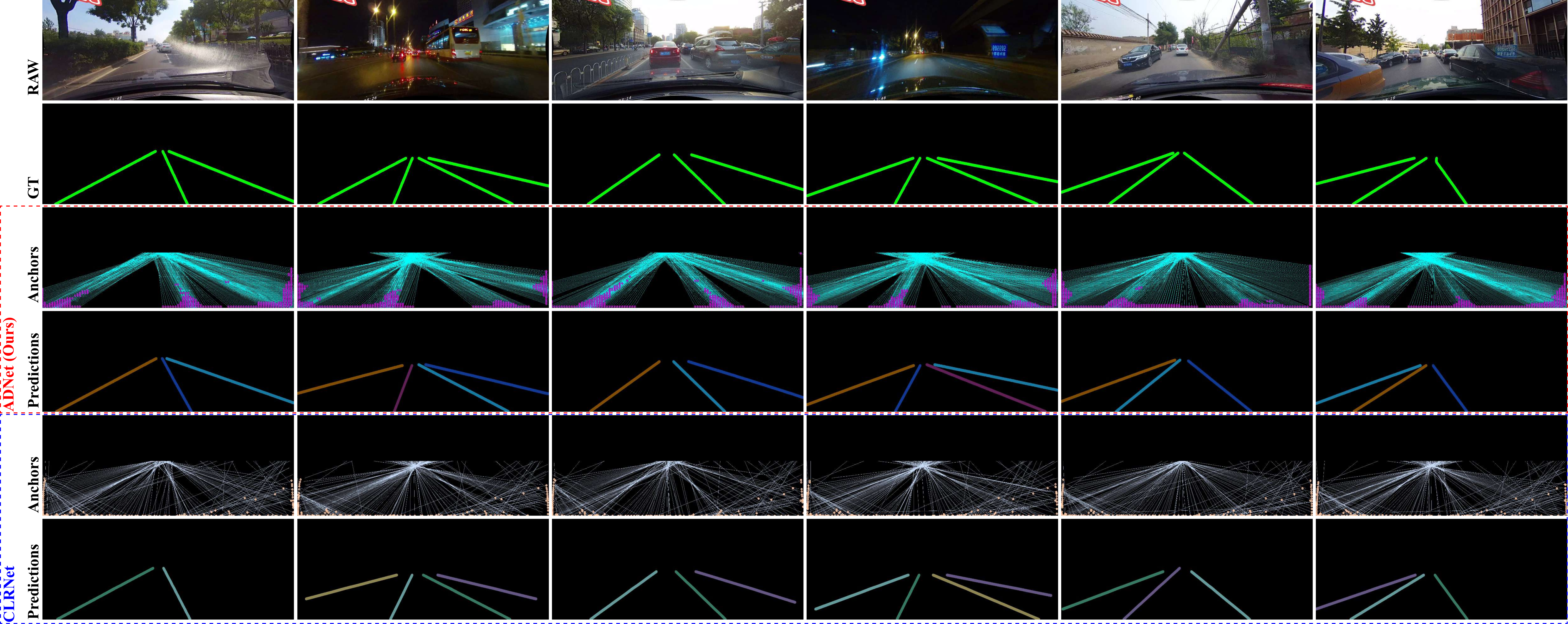}
}
\vspace{-3mm}
\quad

\subfigure[Visualisation results on TuSimple]{
\centering
\includegraphics[width=\linewidth]{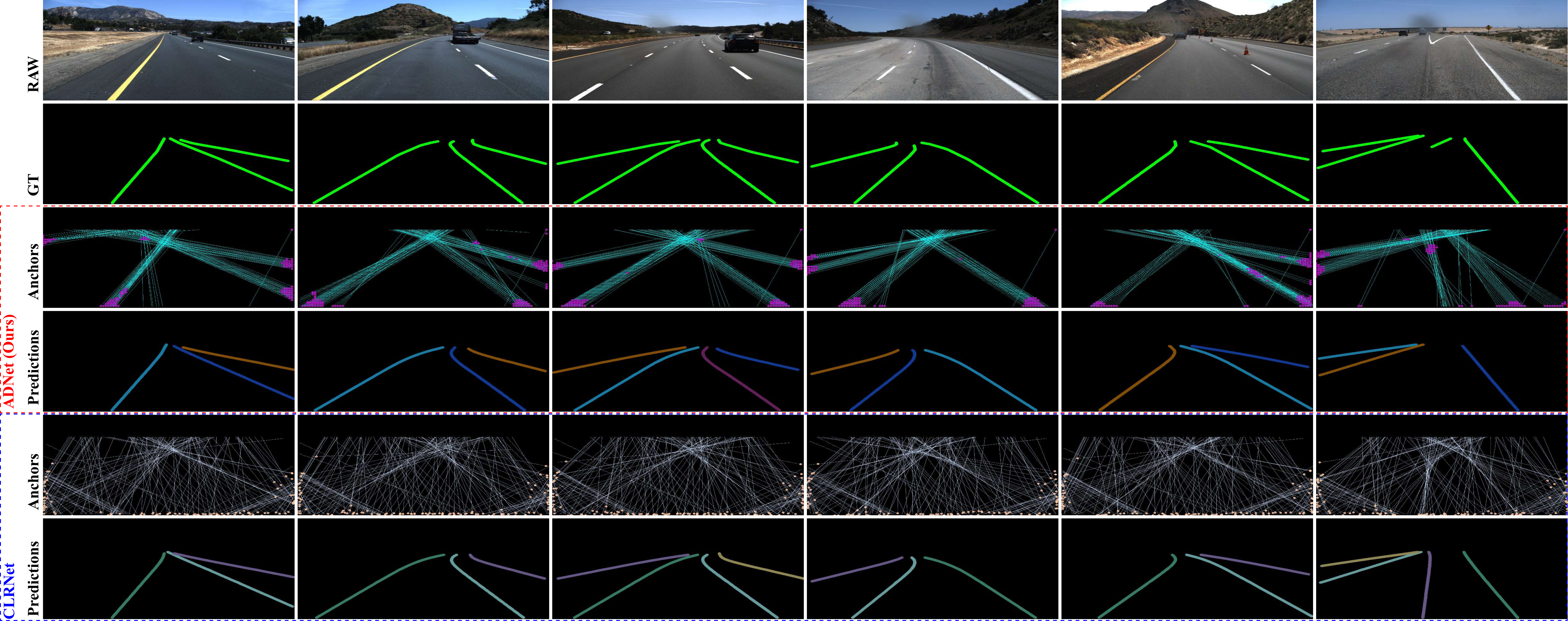}
}
\vspace{-3mm}
\caption{More visualisation results on VIL-100, CULane and TuSimple compared with CLRNet~\cite{Clrnet}. Best view in zoom.}
\label{vis_spp}
\end{figure*}
\end{document}